\documentclass[10pt,twocolumn,letterpaper]{article}

\usepackage{iccv}
\usepackage{times}
\usepackage{epsfig}
\usepackage{graphicx}
\usepackage{amsmath}
\usepackage{amssymb}
\usepackage{makecell}
\usepackage{multirow}
\usepackage{array}
\usepackage{color}
\usepackage{bm}
\usepackage{threeparttable}
\usepackage[linesnumbered,ruled]{algorithm2e}

\newcommand{\urllink}{\fontsize{9pt}{\baselineskip}\selectfont}

\usepackage[pagebackref=true,breaklinks=true,letterpaper=true,colorlinks,bookmarks=false]{hyperref}

\iccvfinalcopy 

\def\iccvPaperID{989} 

\ificcvfinal\fi
\begin{document}

\title{\LaTeX}
\title{DensePoint: Learning Densely Contextual Representation for Efficient Point Cloud Processing}

\author{Yongcheng Liu$^{\dag \hspace{0.5pt} \ddag}$ \quad Bin Fan\thanks{Corresponding author: Bin Fan}\hspace{4pt}$^{\dag}$ \quad Gaofeng Meng$^{\dag}$ \quad Jiwen Lu$^{\S}$ \quad Shiming Xiang$^{\dag \hspace{0.5pt} \ddag}$ \quad Chunhong Pan$^{\dag}$\\
$^{\dag}$\hspace{1pt}National Laboratory of Pattern Recognition, Institute of Automation, Chinese Academy of Sciences\\
$^{\ddag}$\hspace{1pt}School of Artificial Intelligence, University of Chinese Academy of Sciences\\
$^{\S}$\hspace{1pt}Department of Automation, Tsinghua University\\
{\tt\small \{{yongcheng.liu,\;bfan,\;gfmeng,\;smxiang,\;chpan}\}@nlpr.ia.ac.cn \quad\quad lujiwen@tsinghua.edu.cn}
}

\maketitle
\thispagestyle{empty}

\begin{abstract}
\label{abstract}
Point cloud processing is very challenging, as the diverse shapes formed by irregular points are often indistinguishable.
A thorough grasp of the elusive shape requires sufficiently contextual semantic information, yet few works devote to this.
Here we propose DensePoint, a general architecture to learn densely contextual representation for point cloud processing.
Technically, it extends regular grid CNN to irregular point configuration by generalizing a convolution operator, which holds the permutation invariance of points, and achieves efficient inductive learning of local patterns.
Architecturally, it finds inspiration from dense connection mode, to repeatedly aggregate multi-level and multi-scale semantics in a deep hierarchy.
As a result, densely contextual information along with rich semantics, can be acquired by DensePoint in an organic manner, making it highly effective.
Extensive experiments on challenging benchmarks across four tasks, as well as thorough model analysis, verify DensePoint achieves the state of the arts.


\end{abstract}

\section{Introduction}
\label{sec:introduction}
\begin{figure}[t]
\centerline{\includegraphics[width=8cm]{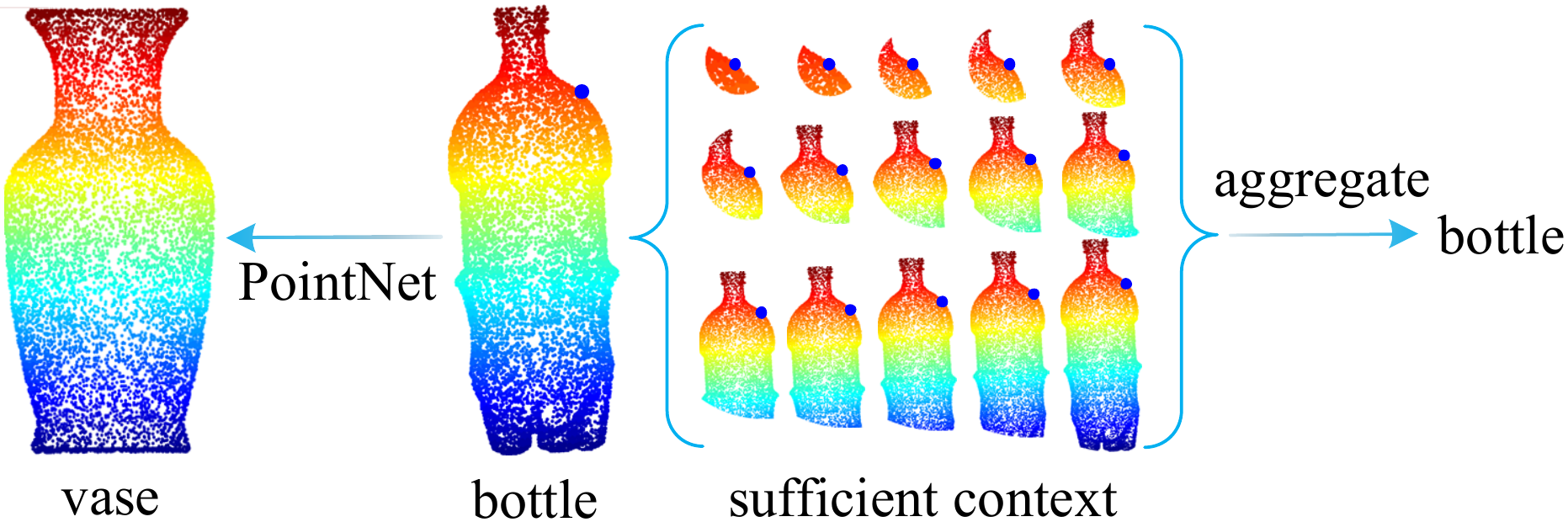}}
\caption{\textbf{Motivation}: sufficiently contextual semantic information is essential for a thorough
grasp of the elusive shape formed by point cloud. The ``bottle'' is misidentified as the ``vase'' by PointNet~\cite{c1_pointnet}, while with sufficient context aggregated, it can be accurately recognized. Here, we only illustrate the multi-level context around the \textcolor[rgb]{0.00,0.00,1.00}{blue} point for visual clearness.}
\label{fig1:motivation}
\vspace{-13pt}
\end{figure}

Recently, the processing of point cloud, which comprises an irregular set of 3D points, has drawn a lot of attention, due to its wide range of applications such as robot manipulation~\cite{robotic} and autonomous driving~\cite{self-drive}.
However, modern applications usually demand for a high-level understanding of point cloud, \textit{i.e.}, identifying the implicit 3D shape pattern.
This is quite challenging, since the diverse shapes, abstractly formed by these irregular points, are often hardly distinguishable.
For this issue, it is essential to capture sufficiently contextual semantic information for a thorough grasp of the elusive shape (see Fig.~\ref{fig1:motivation} for details).

Over the past few years, convolutional neural network (CNN) has demonstrated its powerful abstraction ability of semantic information in image recognition field~\cite{visualize}.
%
Accordingly, much effort is focused on replicating its remarkable success on the analysis of image~\cite{Alexnet,VGG}, \textit{i.e.}, regular grid data, to irregular point cloud processing \cite{c2_pointnet2,c23,c6_synccnn,c14_scn,c69,liu2019rscnn}.
A straightforward strategy is to transform point cloud into regular voxels~\cite{modelnet40,vox2,c45} or multi-view images~\cite{multiview1,c37,multiview2}, for easy application of CNN.
These transformations, however, usually lead to much loss of rich 3D geometric information, as well as high complexity.

Another difficult yet attractive solution is to learn directly from irregular point cloud.
PointNet~\cite{c1_pointnet}, a pioneer in this direction, achieves the permutation invariance of points by learning over each point independently, then applying a symmetric function to accumulate features.
Though impressive, it ignores local patterns that have been proven to be important for abstracting high-level visual semantics in image CNN~\cite{visualize}.
To remedy this defect, KCNet~\cite{c9_kcnet} mines local patterns by creating a k-NN graph over each point in PointNet.
Nevertheless, it inherits another defect of PointNet, \textit{i.e.}, no pooling layer to explicitly raise the level of semantics.
PointNet++~\cite{c2_pointnet2} hierarchically groups point cloud into local subsets and learns on them by PointNet.
This design indeed works like CNN, but the basic operator, PointNet, demands high complexity for enough effectiveness.

Besides high-level semantics, contextual information, which reflects the potential semantic dependencies between a target pattern and its surroundings~\cite{c71}, is also critical for shape pattern recognition.
A typical approach in this view is multi-scale learning.
Accordingly, PointNet++~\cite{c2_pointnet2} directly applies multi-scale grouping in each layer, \textit{i.e.}, capturing context at the same semantic level.
This way, however, is suboptimal as it ignores the inherent difference in semantic levels at different scales, and often causes huge computational cost, especially for lots of scales.
Multi-resolution grouping~\cite{c2_pointnet2} can partly alleviate the latter issue, yet actually, it also abandons crucial context acquisition.
ShapeContextNet~\cite{c7_attsp} finds another strategy inspired by shape context~\cite{shapecontext}.
It applies self-attention~\cite{selfatt} in each layer of PointNet~\cite{c1_pointnet} to dynamically learn the relation weight among all points, and regards this weight as global shape context.
Though fully automatic, it lacks an explicit semantic abstraction like CNN from local to global, and the weight matrix $N\times N$ in self-attention can cause huge complexity when the number of points $N$ increases.

In short, there are mainly two key requirements to exploit CNN for effective learning on point cloud:
%
1) A convolution operator on point cloud, which can be permutation invariant to unordered points, and can achieve efficient inductive learning of local patterns, is required;
2) A deep hierarchy, which can acquire sufficiently contextual semantics for accurate shape recognition, is also required.

\begin{figure*}[t]
\centerline{\includegraphics[width=17cm]{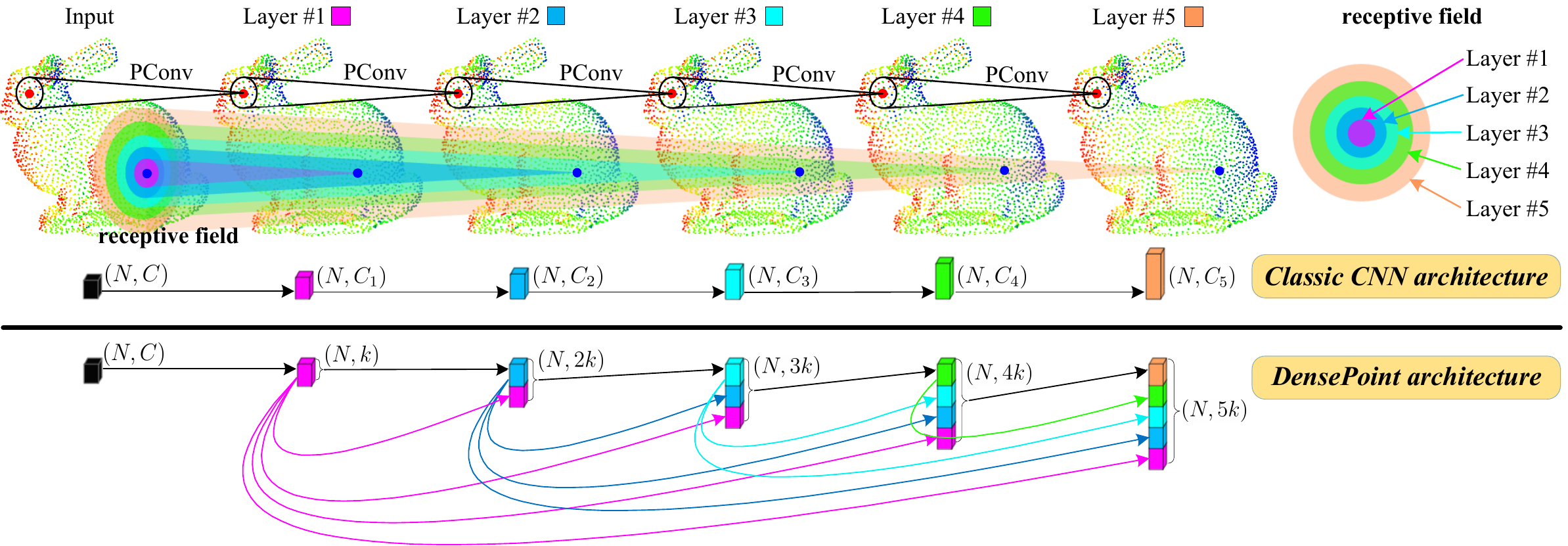}}
\caption{The illustration of DensePoint. It extends regular grid CNN to irregular point configuration by an efficient generalized convolution operator (PConv in Eq.~\eqref{Eq1:general_conv}). Instead of classic CNN architecture with layer-by-layer connections, it finds inspiration from dense connection mode~\cite{densenet}, to repeatedly aggregate multi-level along with multi-scale semantics in an organic manner. To avoid high complexity in deep layers, it forces the output of each layer to be equally narrow with a small constant $k$ (\textit{e.g.}, 24). As a result, densely contextual representation can be learned efficiently for point cloud processing. Here, $N$ is the number of points while $C$, $C_{*}$ and $k$ denote feature dimension.}
\label{fig2:DensePoint}
\vspace{-10pt}
\end{figure*}

Accordingly, we propose DensePoint, a general architecture to learn densely contextual representation for point cloud processing, as illustrated in Fig.~\ref{fig2:DensePoint}.
Technically, DensePoint extends regular grid CNN to irregular point configuration by generalizing a convolution operator, which holds the permutation invariance of points, and respects the convolutional properties, \textit{i.e.}, local connectivity and weight sharing.
Owing to its efficient inductive learning of local patterns, a deep hierarchy can be easily built in DensePoint for semantic abstraction.
Architecturally, DensePoint finds inspiration from dense connection mode~\cite{densenet}, to repeatedly aggregate multi-level and multi-scale semantics in the deep hierarchy.
As a result, densely contextual information along with rich semantics, can be acquired by DensePoint in an organic manner, making it highly effective.

The key contributions are highlighted as follows:
\begin{itemize}
\setlength{\itemsep}{0ex}
\vspace{-5pt}
\item A generalized convolution operator is formulated. It is permutation invariant to points, and respects the convolutional properties of local connectivity and weight sharing, thus extending regular grid CNN to irregular configuration for efficient point cloud processing.
\vspace{-2pt}
\item A general architecture equipped with the generalized convolution operator to learn densely contextual representation of point cloud, \textit{i.e.}, DensePoint, is proposed. It can acquire sufficiently contextual semantic information for accurate recognition of the implicit shape.
\vspace{-13pt}
\item Comprehensive experiments on challenging benchmarks across four tasks, \textit{i.e.}, shape classification, shape retrieval, part segmentation and normal estimation, as well as thorough model analysis, demonstrate that DensePoint achieves the state of the arts.
\end{itemize} 

\section{Related Work}
\label{sec:related work}
In this section, we briefly review existing deep learning methods for 3D shape learning.

\noindent \textbf{View-based and volumetric methods.} \
View-based methods~\cite{multiview1,c37,multiview2,multiview3,c49,c51,c31} represent a 3D shape as a collection of 2D views, over which classic CNN used in image analysis field can be easily applied.
However, 2D projections could cause much loss of 3D shape information due to many self-occlusions. Volumetric methods convert a 3D shape into a regular 3D grid~\cite{modelnet40,vox2,c45}, over which 3D CNN~\cite{C3D} can be employed. The main limitation is the quantization loss of 3D shape information due to the low resolution enforced by 3D grid. Although this issue can be partly rescued by recent space partition methods like K-d trees~\cite{c26} or octrees~\cite{c29,vox3,c53,c70}, they still rely on a subdivision of a bounding volume. By contrast, our work devotes to learn directly from irregular 3D point cloud.

\noindent \textbf{Deep learning on point cloud.} \
Much effort has been focused on learning directly on point cloud. PointNet~\cite{c1_pointnet} pioneers this route by learning on each point independently and accumulating the final features. Yet it ignores local patterns, which limits its semantic learning ability. Accordingly, some works~\cite{c2_pointnet2,PCPNet,c9_kcnet} partition point cloud into local subsets and learn on them based on PointNet. Some other works introduce graph convolutional network to learn over a local graph~\cite{c14_scn,c19,c30} or geometric elements~\cite{c8_superpoint}. However, these methods either lack an explicit semantic abstraction like CNN from local to global, or cause considerable complexity. By contrast, our work extends regular grid CNN to irregular point configuration, achieving efficient learning for point cloud processing.

In addition, there are some works mapping point cloud into a regular space to facilitate the application of classic CNN, \textit{e.g.}, a sparse lattice structure~\cite{c10_splatnet} with bilateral convolution~\cite{bcl} or a continuous volumetric function~\cite{c16_eocnn} with 3D CNN. Nevertheless, in our case, we learn directly from irregular point cloud, which is much more challenging.


\noindent \textbf{Contextual learning on point cloud.} \
Contextual information is important for identifying the implicit shape pattern. PointNet++~\cite{c2_pointnet2} follows the traditional multi-scale learning by directly capturing context on the same layer, which often causes huge complexity. Hence an alternate called multi-resolution grouping~\cite{c2_pointnet2} is devised for efficiency. It forces each layer to learn from its previous layer and the raw input (on the same local region) simultaneously. However, this can be less effective as it actually abandons crucial context acquisition. ShapeContextNet~\cite{c7_attsp} finds another strategy inspired by shape context~\cite{shapecontext}. Instead of the traditional handcrafted design, it applies self-attention~\cite{selfatt} to dynamically learn a weight for all point pairs. Though fully automatic, it lacks a local-to-global semantic learning like CNN. By contrast, we develop a deep hierarchy by an efficient generalized convolution operator, and organically aggregate multi-level contextual semantics in this hierarchy.

\section{Method}
\label{sec:method}
In this section, we first describe the generalized convolution operator and the pooling operator on point cloud. Then, we present DensePoint, and elaborate how it learns densely contextual representation for point cloud processing.

\subsection{Convolution and Pooling on Point Cloud}
\label{sec:pooling}
\noindent\textbf{PConv: convolution on point cloud.} \
Classic convolution on the image operates on a local grid region (\textit{i.e.}, local connectivity), and the convolution filter weights of this grid region are shared along the spatial dimension (\textit{i.e.}, weight sharing). However, this operation is difficult to implement on point cloud due to the irregularity of points. To deal with this problem, we decompose the classic convolution into two core steps, \textit{i.e.}, feature transformation and feature aggregation. Accordingly, a generalized convolution on point cloud can be formulated as
\begin{equation}
\label{Eq1:general_conv}
{\bm{\mathrm f}}_{\mathcal{N}(x)} = \rho \big( \{\phi({\bm{\mathrm f}}_{x_n}), \ \forall x_n \in \mathcal{N}(x)\} \big),
\end{equation}
where both $x$ and $x_n$ denote a 3D point in $\mathbb{R}^3$, and ${\bm{\mathrm f}}$ is feature vector. $\mathcal{N}(x)$, the neighborhood formed by a local point cloud to convolve, is sampled from the whole point cloud by taking a sampled point $x$ as the centroid, and the nearby points as its neighbors $x_n$. ${\bm{\mathrm f}}_{\mathcal{N}(x)}$, the convolutional result as the inductive representation of $\mathcal{N}(x)$, is obtained by: (i) performing a feature transformation with function $\phi$ on each point in $\mathcal{N}(x)$; (ii) applying a aggregation function $\rho$ to aggregate these transformed features. Finally, as shown in the upper part of Fig.~\ref{fig2:DensePoint} (PConv), similar to classic grid convolution, ${\bm{\mathrm f}}_{\mathcal{N}(x)}$ is assigned to be the feature vector of the centroid point $x$ in the next layer. Noticeably, some previous works such as~\cite{c2_pointnet2} also use this general formulation.

In Eq.~\eqref{Eq1:general_conv}, ${\bm{\mathrm f}}_{\mathcal{N}(x)}$ can be permutation invariant only when the inner function $\phi$ is shared over each point in ${\mathcal{N}(x)}$, and the outer function $\rho$ is symmetric (\textit{e.g.}, sum). Accordingly, for high efficiency, we employ a shared single-layer perceptron (SLP, for short) following a nonlinear activator, as $\phi$ to implement feature transformation. Meanwhile, as done in classic convolution, $\phi$ is also shared over each local neighborhood, for achieving the weight sharing mechanism. As a result, with a symmetric $\rho$, the generalized PConv can achieve efficient inductive learning of local patterns, whilst be independent of the irregularity of points. Further, using PConv as the basic operator, a classic CNN architecture (no downsampling), as shown in the upper part of Fig.~\ref{fig2:DensePoint}, can be easily built with layer-by-layer connections.


\noindent\textbf{PPool: pooling on point cloud.} \
In classic CNN, pooling is usually performed to explicitly raise the semantic level of the representation and improve computational efficiency. Here, using PConv, this operation can be achieved on point cloud in a learnable way. Specifically, $N_o$ points are first uniformly sampled from the input $N_i$ points, where $N_o < N_i$ (\textit{e.g.}, $N_o = N_i/2$). Then, PConv can be applied to convolve all the local neighborhoods centered on those $N_o$ points, to generate a new downsampling layer.

\subsection{Learning Densely Contextual Representation}
\noindent\textbf{Classic CNN architecture.} \
\label{Sec:normal_arch}
In a classic CNN architecture with layer-by-layer connections (the upper part of Fig.~\ref{fig2:DensePoint}), hierarchical representations can be learned with the low-level ones in early layers and the high-level ones in deep layers~\cite{visualize}. However, a significant drawback is that each layer can only learn from single-level representation. As a consequence, all layers can capture only single-scale shape information from the input point cloud. Formally, assume a point cloud ${\bm{\mathrm P}}^{0}$ that is passed through this type of network. The network comprises $L$ layers, in which the $\ell^{th}$ layer performs a non-linear transformation $\mathcal{H}^{\ell}(\cdot)$. Then, the output of the $\ell^{th}$ layer can be learned from its previous layer as
\begin{equation}
\label{Eq2:layer_by_layer}
{\bm{\mathrm P}}^{\ell} = \mathcal{H}^{\ell}({\bm{\mathrm P}}^{\ell-1}),
\end{equation}
where each point in ${\bm{\mathrm P}}^{\ell-1}$ is of single-scale receptive filed on the input point cloud ${\bm{\mathrm P}}^{0}$, resulting that the learned ${\bm{\mathrm P}}^{\ell}$ captures only single-scale shape information. Finally, this will lead to a weakly contextual representation, which is not effective enough for identifying the diverse implicit shapes.

\noindent\textbf{DensePoint architecture.} \
To overcome the above issue, we present a general architecture, \textit{i.e.}, DensePoint shown in the lower part of Fig.~\ref{fig2:DensePoint}, inspired by dense connection mode~\cite{densenet}. Specifically, for each layer in DensePoint (no downsampling), the outputs of all preceding layers are used as its input, and its own output is used as the input to all subsequent layers. That is, ${\bm{\mathrm P}}^{\ell}$ in Eq.~\eqref{Eq2:layer_by_layer} becomes
\label{Sec:densepoint}
\begin{equation}
\label{Eq3:dense_connect}
{\bm{\mathrm P}}^{\ell} = \mathcal{H}^{\ell} \big( [{\bm{\mathrm P}}^{0}, {\bm{\mathrm P}}^{1}, \dots, {\bm{\mathrm P}}^{\ell-1}] \big),
\end{equation}
where $[\cdot]$ denotes the concatenation of the outputs of all preceding layers. Here, ${\bm{\mathrm P}}^{\ell}$ is forced to learn from multi-level representations, which facilitates to aggregate multi-level shape semantics along with multi-scale shape information. In this way, each layer in DensePoint can capture a certain level (scale) of context, and the level can be gradually increased as the network deepens. Moreover, the acquired dense context in deep layers can also improve the abstraction of high-level semantics in turn, making the whole learning process organic. Eventually, very rich local-to-global shape information in the input ${\bm{\mathrm P}}^{0}$ can be \textit{progressively} aggregated together, resulting in a densely contextual representation for point cloud processing.

Note that DensePoint is quite different from the traditional multi-scale strategy~\cite{c2_pointnet2}. The former progressively aggregates multi-level (multi-scale) semantics that is organically learned by each layer, while the latter artlessly gathers multi-scale information at the same level. It is also dissimilar to a simple concatenation of all layers as the final output, which results in each layer being less contextual.

\noindent\textbf{Narrow architecture.} \
When the network deepens, DensePoint will suffer from high complexity, since the convolutional overhead of deep layers will be huge with all preceding layers as the input. Thus, we narrow the output channels of each layer in DensePoint with a small constant $k$ (\textit{e.g.}, 24), instead of the large ones (\textit{e.g.}, 512) in classic CNN.

\noindent\textbf{ePConv: enhanced PConv.} \
Though lightweight, such narrow DensePoint will lack the expressive power, since with much narrow output $k$, the shared SLP in PConv, \textit{i.e.}, $\phi$ in Eq.~\eqref{Eq1:general_conv}, could be insufficient in terms of learning ability. To overcome this issue, we introduce the filter grouping~\cite{Alexnet} to enhance PConv, which divides all the filters in a layer into several groups, and each group performs individual operation. Formally, the enhanced PConv (ePConv, for short) converts Eq.~\eqref{Eq1:general_conv} to
\label{Sec:epconv}
\begin{equation}
\label{Eq4:enhance_conv}
{\bm{\mathrm f}}_{\mathcal{N}(x)} = \psi \big( \rho \big( \{\widetilde{\phi}({\bm{\mathrm f}}_{x_n}), \ \forall x_n \in \mathcal{N}(x)\} \big) \big),
\end{equation}
where $\widetilde{\phi}$, the grouped version of SLP $\phi$, can widen its output to enhance its learning ability and maintain the original efficiency, and $\psi$, a normal SLP (shared over each centroid point $x$), is added to integrate the detached information in all groups. Both $\widetilde{\phi}$ and $\psi$ include a nonlinear activator.

\begin{algorithm}[t]
\small
\caption{DensePoint forward pass algorithm}
\label{algo}
\KwIn{point cloud ${\bm{\mathrm P}}$; input features \{${\bm{\mathrm f}}_x, \forall x \in {\bm{\mathrm P}}$\}; depth $L$; weight ${\bm{\mathrm W}}_{\widetilde{\phi}}^{\ell}$, ${\bm{\mathrm W}}_{\psi}^{\ell}$ and bias ${\bm{\mathrm b}}_{\widetilde{\phi}}^{\ell}$, ${\bm{\mathrm b}}_{\psi}^{\ell}$ for SLP $\widetilde{\phi}$ and SLP $\psi$ in Eq.~\eqref{Eq4:enhance_conv}, $\forall \ell \in$ \{1, ..., $L$\}; non-linearity $\sigma$; aggregation function $\rho$; neighborhood method $\mathcal{N}$}
\KwOut{densely contextual representations \{${\bm{\mathrm r}}_x$, $\forall x \in {\bm{\mathrm P}}$\}}
${\bm{\mathrm f}}^{0}_x \leftarrow {\bm{\mathrm f}}_x, \forall x \in {\bm{\mathrm P}}$\;
\For{$\ell=1...L$}
{
    \For{$x \in {\bm{\mathrm P}}$}
    {
        ${\bm{\mathrm f}}^{\ell}_{\mathcal{N}(x)} \leftarrow \rho \big( \{ \sigma( \widetilde{\bm{\mathrm W}}_{\widetilde{\phi}}^{\ell} \cdot \widetilde{\bm{\mathrm f}}^{\ell-1}_{x_n} + \widetilde{\bm{\mathrm b}}_{\widetilde{\phi}}^{\ell} ), \ \forall x_n \in \mathcal{N}(x) \} \big)$\;
        ${\bm{\mathrm f}}^{\ell}_{x} \leftarrow \sigma ( {\bm{\mathrm W}}_{\psi}^{\ell} \cdot {\bm{\mathrm f}}^{\ell}_{\mathcal{N}(x)} + {\bm{\mathrm b}}_{\psi}^{\ell} )$\;
    }
    ${\bm{\mathrm f}}^{\ell}_{x} \leftarrow [ {\bm{\mathrm f}}^{0}_x, ..., {\bm{\mathrm f}}^{\ell}_x], \ \forall x \in {\bm{\mathrm P}}$\;
}
return \{${\bm{\mathrm r}}_x \leftarrow {\bm{\mathrm f}}^{L}_{x}, \ \forall x \in {\bm{\mathrm P}}$\}
\footnotetext{No sibling is visited}
\end{algorithm}

To elaborate ePConv with filter grouping, let SLP$^{\widetilde{\phi}}$ (resp. SLP$^{\psi}$) denote the SLP of $\widetilde{\phi}$ (resp. $\psi$), and $C_i$ (resp. $C_o$) denote the input (resp. output) channels of SLP$^{\widetilde{\phi}}$. $N_g$ is the number of groups. Then, the parameter number of SLP$^{\widetilde{\phi}}$ before and after filter grouping is, $C_i \times C_o$ vs. $(C_i/N_g) \times (C_o/N_g) \times N_g = (C_i \times C_o)/N_g$. Here $C_i$ and $C_o$ are divisible by $N_g$ and the few parameters in the bias term are ignored for clearness. In other words, using filter grouping, $C_o$ can be increased by $N_g$ times but with almost the same complexity. Besides, inspired by the bottleneck layer~\cite{resnet}, we fix the output channels of SLP$^{\widetilde{\phi}}$ and SLP$^{\psi}$ as $C_o:k=4:1$ (\textit{i.e.}, $C_o=4k$), to hold the original narrowness for DensePoint. Hence, with a small $k$, SLP$^{\psi}$ actually leads to only a little complexity of $4k\times k$, which can be easily remedied by a suitable $N_g$. The detailed forward pass procedure of DensePoint equipped with ePConv can be referred in Algorithm~\ref{algo}, where $\widetilde{*}$ indicates performing grouping operation.

\begin{figure}[t]
\centerline{\includegraphics[width=8.2cm]{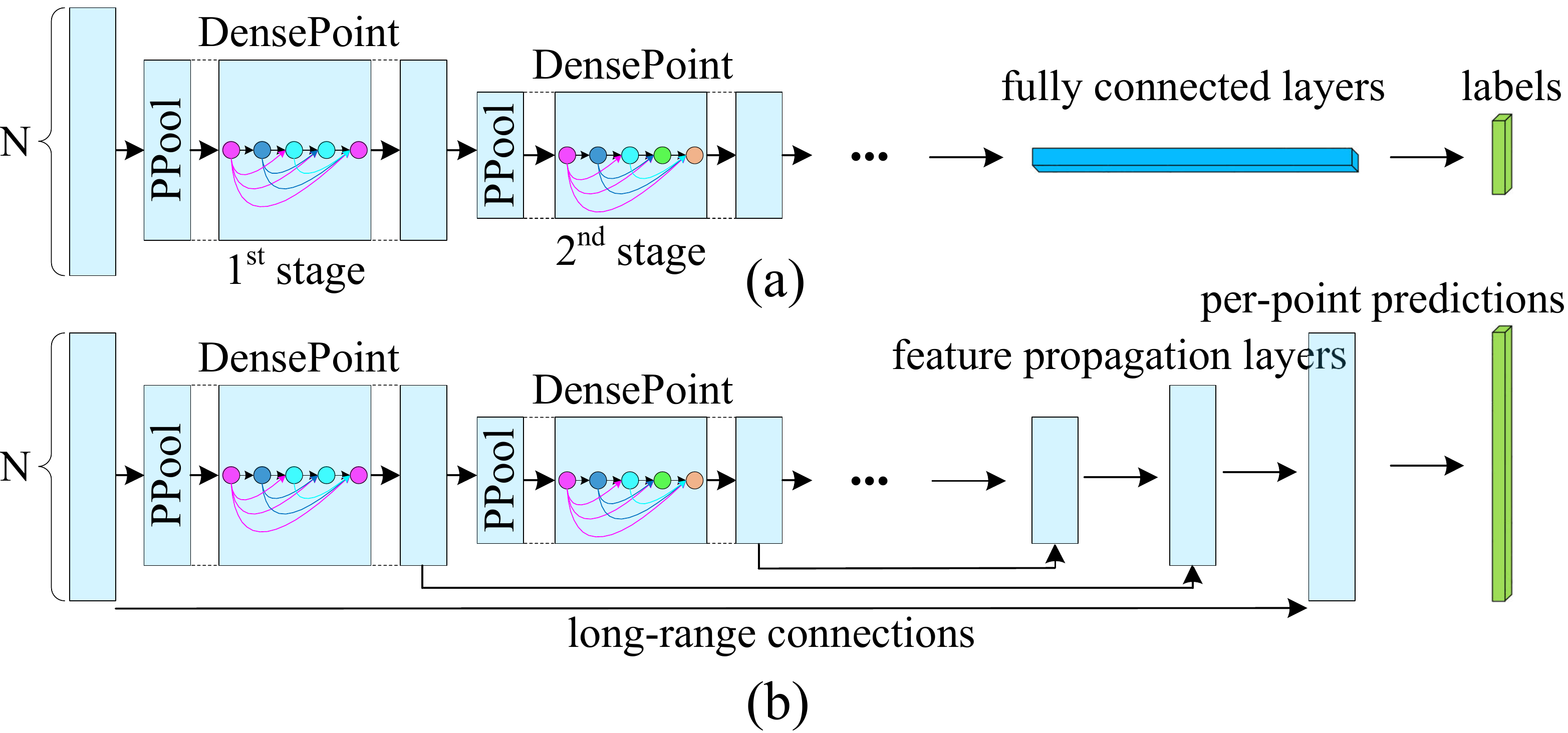}}
\caption{DensePoint applied in point cloud classification (a) and per-point analysis (b). PPool: pooling on point cloud (Sec~\ref{sec:pooling}). $N$ is the number of points. The stage means several successive layers with the same number of points.}
\label{fig3:arch}
\vspace{-12pt}
\end{figure}

\noindent\textbf{DensePoint for point cloud processing.} \
DensePoint applied in point cloud classification and per-point analysis (\textit{e.g.}, segmentation) are illustrated in Fig.~\ref{fig3:arch}. In both tasks, DensePoint with ePConv is applied in each stage of the network to learn densely contextual representation, while PPool with original PConv is used to explicitly raise the semantic level and improve efficiency. For classification, the final global representation is learned by three PPools and two DensePoints (11 layers in total, $L=11$), followed by three fully connected (fc) layers as the classifier. For per-point analysis, four levels of representations learned by four PPools and three DensePoints (17 layers in total, $L=17$), are sequentially upsampled by feature propagation~\cite{c2_pointnet2} to generate per-point predictions. All the networks can be trained in an end-to-end manner. The configuration details are included in the supplementary material.

\noindent\textbf{Implementation details.} \
PPool: the farthest points are picked from point cloud for uniform downsampling. Neighborhood: the spherical neighborhood is adopted; a fixed number of neighbors are randomly sampled in each neighborhood for batch processing (the centroid is reused if not enough), and they are normalized by subtracting the centroid. Group number $N_g$ in $\widetilde{\phi}$ (Eq.~\eqref{Eq4:enhance_conv}): $N_g = 2$. Nonlinear activator: ReLU~\cite{relu}. Dropout~\cite{dropout}: for model regularization, we apply dropout with 20\% ratio on ${\bm{\mathrm f}}_{\mathcal{N}(x)}$ in Eq.~\eqref{Eq4:enhance_conv} and dropout with 50\% ratio on the first two fc layers in the classification network (Fig.~\ref{fig3:arch}(a)). Narrowness $k$: $k=24$. Aggregation function $\rho$: symmetric function max pooling is employed. Batch normalization (BN)~\cite{BN}: as done in image CNN, BN is used before each nonlinear activator for all layers. Note that only 3D coordinates ($X,Y,Z$) in $\mathbb{R}^{3}$ are used as the initial input features. Code is available\footnote{\urllink\texttt{https://github.com/Yochengliu/DensePoint}}.


\section{Experiment}
\label{sec:experiment}
We conduct comprehensive experiments to validate the effectiveness of DensePoint. We first evaluate DensePoint for point cloud processing on challenging benchmarks across four tasks (Sec~\ref{subsec4.1}). We then provide detailed experiments to study DensePoint thoroughly (Sec~\ref{subsec4.2}).

\subsection{DensePoint for Point Cloud Processing}
\label{subsec4.1}
\noindent \textbf{Shape classification.} \
We evaluate DensePoint on ModelNet40 and ModelNet10 classification benchmarks~\cite{modelnet40}. The former comprises 9843 training models and 2468 test models in 40 classes, while the latter consists of 3991 training models and 908 test models in 10 classes. The point cloud data is sampled from these models by~\cite{c1_pointnet}. For training, we uniformly sample 1024 points as the input. As in~\cite{c26}, we augment the input with random anisotropic scaling in range [-0.66, 1.5] and translation in range [-0.2, 0.2]. For testing, similar to~\cite{c1_pointnet,c2_pointnet2}, we apply voting with 10 tests using random scaling and then average the predictions.

The quantitative comparisons with the state-of-the-art point-based methods are summarized in Table~\ref{Tab1:cls}. Our DensePoint outperforms all the point-input methods. Specifically, it reduces the error rate of PointNet++ by 26.9\% on ModelNet40, and also surpasses its advanced version that applies additional normal data with very dense points (5k). Furthermore, even using only point as the input, DensePoint can also surpass the best additional-input method SO-Net~\cite{c20} by 0.9\% on ModelNet10. These results convincingly verify the effectiveness of DensePoint.

\begin{table}[t]
  \centering
  \small
  \caption{Shape classification results (\textit{overall accuracy}, \%) on ModelNet40 (M40) and ModelNet10 (M10) benchmarks (pnt: point coordinates, nor: normal, ``-'': unknown).}
  \begin{tabular}{lcccc}
  \Xhline{0.8pt}
  method & input & \#points & M40 & M10 \\
  \Xhline{0.5pt}
  Pointwise-CNN~\cite{c17_pointwise} & pnt & 1k & 86.1 & - \\
  Deep Sets~\cite{c24} & pnt & 1k & 87.1 & - \\
  ECC~\cite{c32} & pnt & 1k & 87.4 & 90.8 \\
  PointNet~\cite{c1_pointnet} & pnt & 1k & 89.2 & - \\
  SCN~\cite{c7_attsp} & pnt & 1k & 90.0 & - \\
  Kd-Net(depth=10)~\cite{c26} & pnt & 1k & 90.6 & 93.3 \\
  PointNet++~\cite{c2_pointnet2} & pnt & 1k & 90.7 & - \\
  MC-Conv~\cite{c67} & pnt & 1k & 90.9 & - \\
  KCNet~\cite{c9_kcnet} & pnt & 1k & 91.0 & 94.4 \\
  MRTNet~\cite{c45} & pnt & 1k & 91.2 & - \\
  Spec-GCN~\cite{c14_scn} & pnt & 1k & 91.5 & - \\
  DGCNN~\cite{c22-dgcnn} & pnt & 1k & 92.2 & - \\
  PointCNN~\cite{c27} & pnt & 1k & 92.2 & - \\
  PCNN~\cite{c16_eocnn} & pnt & 1k & 92.3 & 94.9 \\
  \textbf{Ours} & \textbf{pnt} & \textbf{1k} & \textbf{93.2} & \textbf{96.6} \\
  SO-Net~\cite{c20} & pnt & 2k & 90.9 & 94.1 \\
  Kd-Net(depth=15)~\cite{c26} & pnt & 32k & 91.8 & 94.0 \\
  \Xhline{0.5pt}
  O-CNN~\cite{c29} & pnt, nor & - & 90.6 & - \\
  Spec-GCN~\cite{c14_scn} & pnt, nor & 1k & 91.8 & - \\
  PointNet++~\cite{c2_pointnet2} & pnt, nor & 5k & 91.9 & - \\
  SpiderCNN~\cite{c21} & pnt, nor & 5k & 92.4 & - \\
  SO-Net~\cite{c20} & pnt, nor & 5k & 93.4 & 95.7 \\
  \Xhline{0.8pt}
  \end{tabular}
  \label{Tab1:cls}
\end{table}

\begin{table}[t]
  \centering
  \footnotesize
  \caption{Shape retrieval results (mAP, \%) on ModelNet40 (M40) and ModelNet10 (M10) benchmarks (``-'': unknown).}
  \begin{tabular}{llccc}
  \Xhline{0.8pt}
  input & method & \#points/views & M40 & M10 \\
  \Xhline{0.5pt}
  \multirow{2}{*}{Points} & PointNet~\cite{triplet-loss} & 1k & 70.5 & - \\
                          & DGCNN~\cite{c22-dgcnn} & 1k & 85.3 & - \\
                          & PointCNN~\cite{c27} & 1k & 83.8 & - \\
                          & \textbf{Ours} & \textbf{1k} & \textbf{88.5} & \textbf{93.2} \\
  \Xhline{0.5pt}
  \multirow{4}{*}{Images} & GVCNN~\cite{c37} & 12 & 85.7 & - \\
                          & Triplet-center~\cite{triplet-loss} & 12 & 88.0 & - \\
                          & PANORAMA-ENN~\cite{PANORAMA} & - & 86.3 & 93.3 \\
                          & SeqViews~\cite{c49} & 12 & 89.1 & 89.5 \\

  \Xhline{0.8pt}
  \end{tabular}
  \label{Tab2:retrieval}
  \vspace{-12pt}
\end{table}

\begin{table*}[t]
  \centering
  \caption{Shape part segmentation results (\%) on ShapeNet part benchmark (nor: normal, ``-'': unknown).}
  \footnotesize
  \begin{tabular}{p{1.8cm}|p{0.7cm}|p{0.5cm}|p{0.75cm}|p{0.32cm}p{0.32cm}p{0.32cm}p{0.32cm}p{0.32cm}p{0.32cm}p{0.32cm}p{0.32cm}p{0.32cm}p{0.32cm}p{0.32cm}p{0.32cm}p{0.32cm}p{0.32cm}p{0.32cm}p{0.32cm}}
  \Xhline{0.8pt}
  method & input & class mIoU & instance mIoU & air plane & bag & cap & car & chair & ear phone & guitar & knife & lamp & laptop & motor bike & mug & pistol & rocket & skate board & table \\
  \Xhline{0.5pt}
  Kd-Net~\cite{c26} & 4k & 77.4 & 82.3 & 80.1 & 74.6 & 74.3 & 70.3 & 88.6 & 73.5 & 90.2 & 87.2 & 81.0 & 94.9 & 57.4 & 86.7 & 78.1 & 51.8 & 69.9 & 80.3 \\
  PointNet~\cite{c1_pointnet} & 2k & 80.4 & 83.7 & 83.4 & 78.7 & 82.5 & 74.9 & 89.6 & 73.0 & 91.5 & 85.9 & 80.8 & 95.3 & 65.2 & 93.0 & 81.2 & 57.9 & 72.8 & 80.6 \\
  SCN~\cite{c7_attsp} & 1k & 81.8 & 84.6 & 83.8 & 80.8 & 83.5 & 79.3 & 90.5 & 69.8 & \textbf{91.7} & 86.5 & 82.9 & 96.0 & 69.2 & 93.8 & 82.5 & 62.9 & 74.4 & 80.8 \\
  SPLATNet~\cite{c10_splatnet} & - & 82.0 & 84.6 & 81.9 & 83.9 & 88.6 & \textbf{79.5} & 90.1 & 73.5 & 91.3 & 84.7 & 84.5 & \textbf{96.3} & 69.7 & \textbf{95.0} & 81.7 & 59.2 & 70.4 & 81.3 \\
  KCNet~\cite{c9_kcnet} & 2k & 82.2 & 84.7 & 82.8 & 81.5 & 86.4 & 77.6 & 90.3 & 76.8 & 91.0 & 87.2 & 84.5 & 95.5 & 69.2 & 94.4 & 81.6 & 60.1 & 75.2 & 81.3 \\
  RS-Net~\cite{C28} & - & 81.4 & 84.9 & 82.7 & \textbf{86.4} & 84.1 & 78.2 & 90.4 & 69.3 & 91.4 & 87.0 & 83.5 & 95.4 & 66.0 & 92.6 & 81.8 & 56.1 & 75.8 & 82.2 \\
  DGCNN~\cite{c22-dgcnn} & 2k & 82.3 & 85.1 & \textbf{84.2} & 83.7 & 84.4 & 77.1 & 90.9 & 78.5 & 91.5 & 87.3 & 82.9 & 96.0 & 67.8 & 93.3 & 82.6 & 59.7 & 75.5 & 82.0 \\
  PCNN~\cite{c16_eocnn} & 2k & 81.8 & 85.1 & 82.4 & 80.1 & 85.5 & \textbf{79.5} & 90.8 & 73.2 & 91.3 & 86.0 & \textbf{85.0} & 95.7 & 73.2 & 94.8 & \textbf{83.3} & 51.0 & 75.0 & 81.8 \\
  \textbf{Ours} & \textbf{2k} & \textbf{84.2} & \textbf{86.4} & 84.0 & 85.4 & \textbf{90.0} & 79.2 & \textbf{91.1} & \textbf{81.6} & 91.5 & \textbf{87.5} & 84.7 & 95.9 & \textbf{74.3} & 94.6 & 82.9 & \textbf{64.6} & \textbf{76.8} & \textbf{83.7} \\
  \Xhline{0.5pt}
  SO-Net~\cite{c20} & -,nor & 80.8 & 84.6 & 81.9 & 83.5 & 84.8 & 78.1 & 90.8 & 72.2 & 90.1 & 83.6 & 82.3 & 95.2 & 69.3 & 94.2 & 80.0 & 51.6 & 72.1 & 82.6 \\
  SyncCNN~\cite{c3_synccnn} & mesh & 82.0 & 84.7 & 81.6 & 81.7 & 81.9 & 75.2 & 90.2 & 74.9 & 93.0 & 86.1 & 84.7 & 95.6 & 66.7 & 92.7 & 81.6 & 60.6 & 82.9 & 82.1 \\
  \scriptsize{PointNet++}~\cite{c2_pointnet2} & 2k,nor & 81.9 & 85.1 & 82.4 & 79.0 & 87.7 & 77.3 & 90.8 & 71.8 & 91.0 & 85.9 & 83.7 & 95.3 & 71.6 & 94.1 & 81.3 & 58.7 & 76.4 & 82.6 \\
  \scriptsize{SpiderCNN}~\cite{c21} & 2k,nor & 82.4 & 85.3 & 83.5 & 81.0 & 87.2 & 77.5 & 90.7 & 76.8 & 91.1 & 87.3 & 83.3 & 95.8 & 70.2 & 93.5 & 82.7 & 59.7 & 75.8 & 82.8 \\
  \Xhline{0.8pt}
  \end{tabular}
  \label{Tab3:seg}
  \vspace{-5pt}
\end{table*}

\begin{figure}[t]
\centerline{\includegraphics[width=8.2cm]{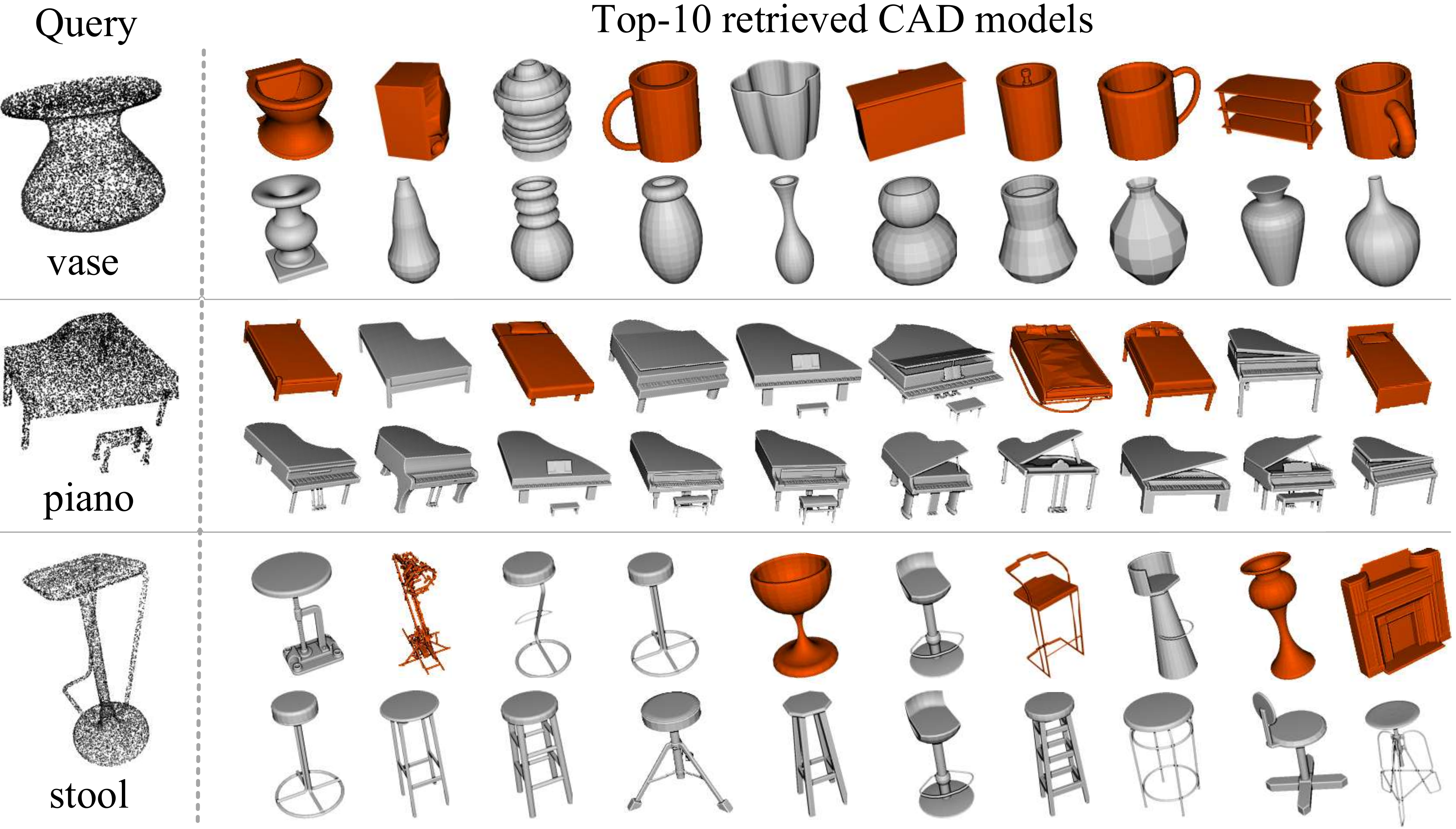}}
\caption{Retrieval examples on ModelNet40. Top-10 matches are shown for each query, with the 1\textsuperscript{st} line for PointNet~\cite{c1_pointnet} and the 2\textsuperscript{nd} line for our DensePoint. The mistakes are highlighted in \textcolor[rgb]{1.00,0.00,0.00}{red}.}
\label{fig4:retrieval}

\end{figure}

\noindent \textbf{Shape retrieval.} \
To further explore the recognition ability of DensePoint for the implicit shapes, we apply the global features, \textit{i.e.}, the outputs of the penultimate fc layer in the classification network (Fig.~\ref{fig3:arch}(a)), for shape retrieval. We sort the most relevant shapes for each query from the test set by cosine distance, and report \textit{mean Average Precision} (mAP). Except for point-based methods, we also compare with some advanced 2D image-based ones. The results are summarized in Table~\ref{Tab2:retrieval}. As can be seen, DensePoint significantly outperforms PointNet by 18\%. It is also comparable with those image-based methods (even the ensemble one~\cite{PANORAMA}), which greatly benefit from image CNN and pre-training with large-scale datasets (\textit{e.g.}, ImageNet~\cite{ImageNet}). Fig.~\ref{fig4:retrieval} shows some retrieval examples.

\noindent \textbf{Shape part segmentation.} \ Part segmentation is a challenging task for fine-grained shape recognition. Here we evaluate DensePoint on ShapeNet part benchmark~\cite{c54}. It contains 16881 shapes with 16 categories, and is labeled in 50 parts in total, where each shape has 2$\sim$5 parts. We follow the data split in~\cite{c1_pointnet}, and similarly, we also randomly pick 2048 points as the input and concatenate the one-hot encoding of the object label to the last feature layer of the segmentation network in Fig.~\ref{fig3:arch}(b). In testing, we also apply voting with ten tests using random scaling. Except for standard IoU (\textit{Inter-over-Union}) score for each category, two types of mean IoU (mIoU) that are averaged across all classes and all instances respectively, are also reported.

\begin{figure}[t]
\centerline{\includegraphics[width=8.2cm]{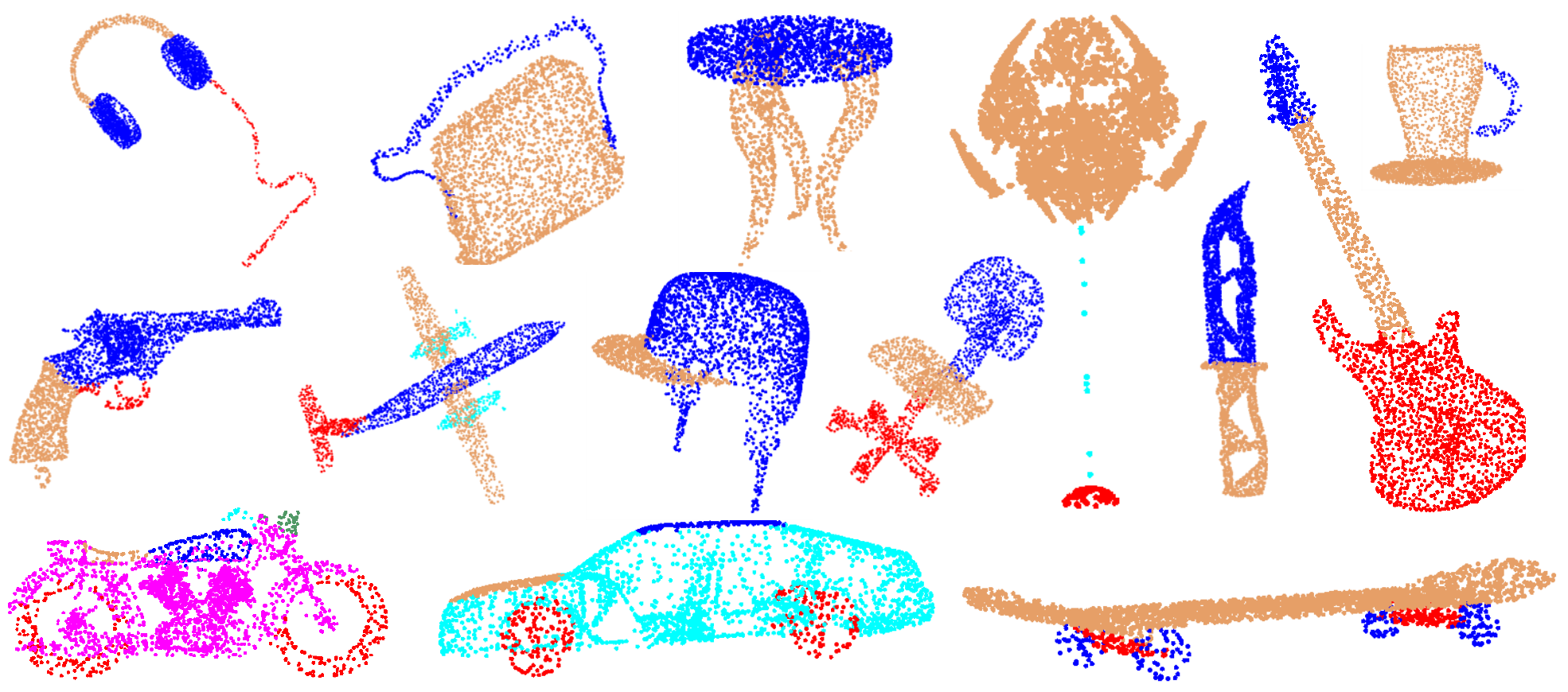}}
\caption{Segmentation examples on ShapeNet part benchmark.}
\label{fig5:part}
\end{figure}

Table~\ref{Tab3:seg} summarizes the quantitative comparisons with the state-of-the-art methods, where DensePoint achieves the best performance. Furthermore, it significantly surpasses the second best point-input methods, \textit{i.e.}, DGCNN~\cite{c22-dgcnn}, with 1.9$\uparrow$ in class mIoU and 1.3$\uparrow$ in instance mIoU respectively. Noticeably, it also sets new state of the arts over the point-based methods in eight categories. These improvements demonstrate the robustness of DensePoint to diverse shapes. Some segmentation examples are shown in Fig.~\ref{fig5:part}.

\begin{table}[t]
  \centering
  \small
  \caption{Normal estimation error on ModelNet40 benchmark.}
  \begin{tabular}{l|lcc}
  \Xhline{0.8pt}
  dataset & method & \#points & error \\
  \Xhline{0.5pt}
  ModelNet40 & PointNet~\cite{c16_eocnn} & 1k & 0.47 \\
                     & PointNet++~\cite{c16_eocnn} & 1k & 0.29 \\
                     & PCNN~\cite{c16_eocnn} & 1k & 0.19 \\
                     & MC-Conv~\cite{c67} & 1k & 0.16 \\
                     & \textbf{Ours} & \textbf{1k} & \textbf{0.149} \\
  \Xhline{0.8pt}
  \end{tabular}
  \label{Tab4:normal}
\end{table}

\noindent \textbf{Normal estimation.} \ Normal estimation in point cloud is a crucial step for numerous applications, from surface reconstruction and scene understanding to rendering. Here, we regard normal estimation as a supervised regression task, and implement it by deploying DensePoint with the segmentation network in Fig.~\ref{fig3:arch}(b). The cosine-loss between the normalized output and the normal ground truth is employed for training. We evaluate DensePoint on ModelNet40 benchmark for this task, where 1024 points are uniformly sampled as the input.

\begin{figure}[t]
\centerline{\includegraphics[width=8.2cm]{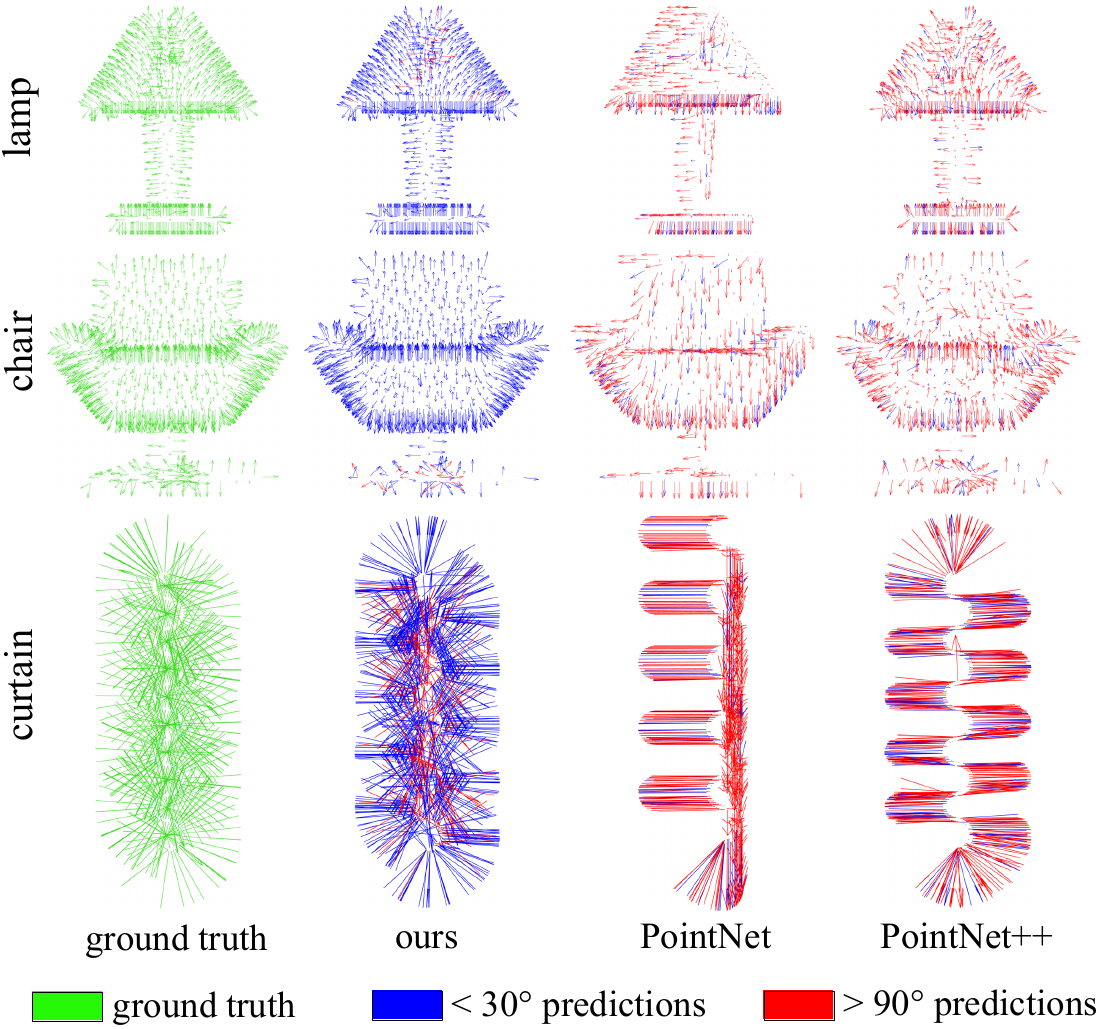}}
\caption{Normal estimation examples on ModelNet40 benchmark. For visual clearness, we only show the predictions with the angle less than 30$^{\circ}$ in blue, and the angle greater than 90$^{\circ}$ in red between the ground truth.}
\label{fig6:normal}
\end{figure}

The quantitative comparisons of the estimation error are summarized in Table~\ref{Tab4:normal}, where DensePoint outperforms other advanced methods. Moreover, it significantly reduces the error of PointNet++ by 48.6\%. Fig.~\ref{fig6:normal} shows some normal prediction examples. As can be seen, DensePoint with densely contextual semantics can obtain more decent normal predictions, while PointNet and PointNet++ present a lot of deviations above 90$^{\circ}$ from the ground truth. However, in this task, DensePoint can not process some intricate shapes well, \textit{e.g.}, curtains and plants.

\subsection{DensePoint Analysis}
\label{subsec4.2}
In this section, we first perform a detailed ablation study for DensePoint. Then, we discuss the group number $N_{g}$ in ePConv (Eq.~\eqref{Eq4:enhance_conv}), the network narrowness $k$, the aggregation function $\rho$ and the network stage to apply DensePoint, respectively. Finally, we analyze the robustness of DensePoint on sampling density and random noise, and investigate the model complexity. All the experiments are conducted on ModelNet40~\cite{modelnet40} dataset.

\vspace{10pt}
\noindent \textbf{Ablation study.} \
The results of ablation study are summarized in Table~\ref{Tab5:ablation}. We set two baselines: model A and model $\overline{\text{A}}$. Model A is set as a classic hierarchical version (the upper part of Fig.~\ref{fig2:DensePoint}, \textit{i.e.}, layer-by-layer connections without contextual learning by DensePoint) of the classification network with the same number of layers, and each layer is configured with the same width. Model $\overline{\text{A}}$ directly concatenates all layers in each stage of model A as the final output of that stage. Both of them are equipped with PConv in Eq.~\eqref{Eq1:general_conv}.

The baseline model A gets a low classification accuracy of 88.6\%, and it increases by only 0.5 percent with direct concatenation (model $\overline{\text{A}}$). However, with the densely contextual semantics of DensePoint, the accuracy raises significantly by 2.5 percent (91.1\%, model B). This convincingly verifies its effectiveness. Then, when using ePConv to enhance the expressive power of each layer in DensePoint, the accuracy can be further improved to 92.5\% (model C). Noticeably, the dropout on ${\bm{\mathrm f}}_{\mathcal{N}(x)}$ in Eq.~\eqref{Eq4:enhance_conv} can bring a boost of 0.3 percent (model D). The data augmentation technique can result in an accuracy variation of 0.7 percent (model E). Finally, by voting strategy, our final model F can achieve an impressive accuracy of 93.2\%. In addition, we also investigate the number of input points by increasing it to 2k, yet obtaining no gain (model G). Maybe the model needs to be modified to adapt for more input points.

\begin{table}[t]
  \centering
  \footnotesize
  \caption{Ablation study of DensePoint (\%) (DA: data augmentation, DP: DensePoint, DO: dropout on ${\bm{\mathrm f}}_{\mathcal{N}(x)}$ in Eq.~\eqref{Eq4:enhance_conv}).}
  \begin{tabular}{cccccccc}
  \Xhline{0.8pt}
  model & \#points & DA & DP & ePConv & DO & vote & acc.\\
  \Xhline{0.5pt}
  A & 1k & $\checkmark$ &  &  &  &  & 88.6 \\
  $\overline{\text{A}}$ & 1k & $\checkmark$ &  &  &  &  & 89.1 \\
  \Xhline{0.5pt}
  B & 1k & $\checkmark$ & $\checkmark$ &  &  &  & 91.1 \\
  C & 1k & $\checkmark$ & $\checkmark$ & $\checkmark$ &  &  & 92.5 \\
  D & 1k & $\checkmark$ & $\checkmark$ & $\checkmark$ & $\checkmark$ &  & 92.8 \\
  E & 1k & $ $ & $\checkmark$ & $\checkmark$ & $\checkmark$ & $ $ & 92.1 \\
  F & 1k & $\checkmark$ & $\checkmark$ & $\checkmark$ & $\checkmark$ & $\checkmark$ & \textbf{93.2} \\
  G & 2k & $\checkmark$ & $\checkmark$ & $\checkmark$ & $\checkmark$ & $\checkmark$ & \textbf{93.2} \\
  \Xhline{0.8pt}
  \end{tabular}
  \label{Tab5:ablation}
\end{table}

\begin{table}[t]
  \centering
  \small
  \caption{The impact of the group number $N_g$ on network parameters, FLOPs and performance ($k=24$).}
  \begin{tabular}{cccc}
  \Xhline{0.8pt}
   group number $N_g$ & \#params & \#FLOPs/sample & acc. (\%) \\
  \Xhline{0.5pt}
   1 & 0.73M & 1030M & 92.7 \\
   2 & 0.67M & 651M & \textbf{93.2} \\
   4 & 0.62M & 457M & 92.2 \\
   6 & 0.61M & 394M & 92.3 \\
  12 & 0.60M & 331M & 92.1 \\
  \Xhline{0.8pt}
  \end{tabular}
  \label{Tab6:grouping}
\end{table}

\vspace{10pt}
\noindent \textbf{Group number $N_g$ in ePConv (Eq.~\eqref{Eq4:enhance_conv}).} \
The filter grouping can greatly reduce the model complexity, whilst leading to a model regularization by rarefying the filter relationships~\cite{Grouping}. Table~\ref{Tab6:grouping} summarizes the impact of $N_g$ on model parameters, model FLOPs (floating point operations/sample) and classification accuracy. As can be seen, the model parameters are very few (0.73M), even though the filter grouping is not performed. This is due to the narrow design ($k=24$) of each layer in DensePoint and few parameters in the generalized convolution operator, ePConv. Eventually, with $N_g=2$, DensePoint can achieve the best result of 93.2\% with acceptable model complexity.

\vspace{10pt}
\noindent \textbf{Network narrowness $k$.} \
Table~\ref{Tab7:narrowness} summarizes the comparisons of different $k$. One can see that a very small DensePoint, \textit{i.e.}, $k=12$, can even obtain an impressive accuracy of 92.1\%. This further verifies the powerfulness of the densely contextual semantics acquired by DensePoint on shape identification. Note that a large $k$ is usually unnecessary for DensePoint, as it will greatly raise the model complexity but not bring any gains.

\vspace{10pt}
\noindent \textbf{Aggregation function $\rho$.} \
We experiment with three symmetric functions, \textit{i.e.}, sum, average pooling and max pooling, whose results are 91.0\%, 91.3\% and 93.2\%, respectively. The max pooling performs best, probably because it can select the biggest feature response to keep the most expressive representation and remove redundant information.

\begin{figure}[t]
\centerline{\includegraphics[width=8.2cm]{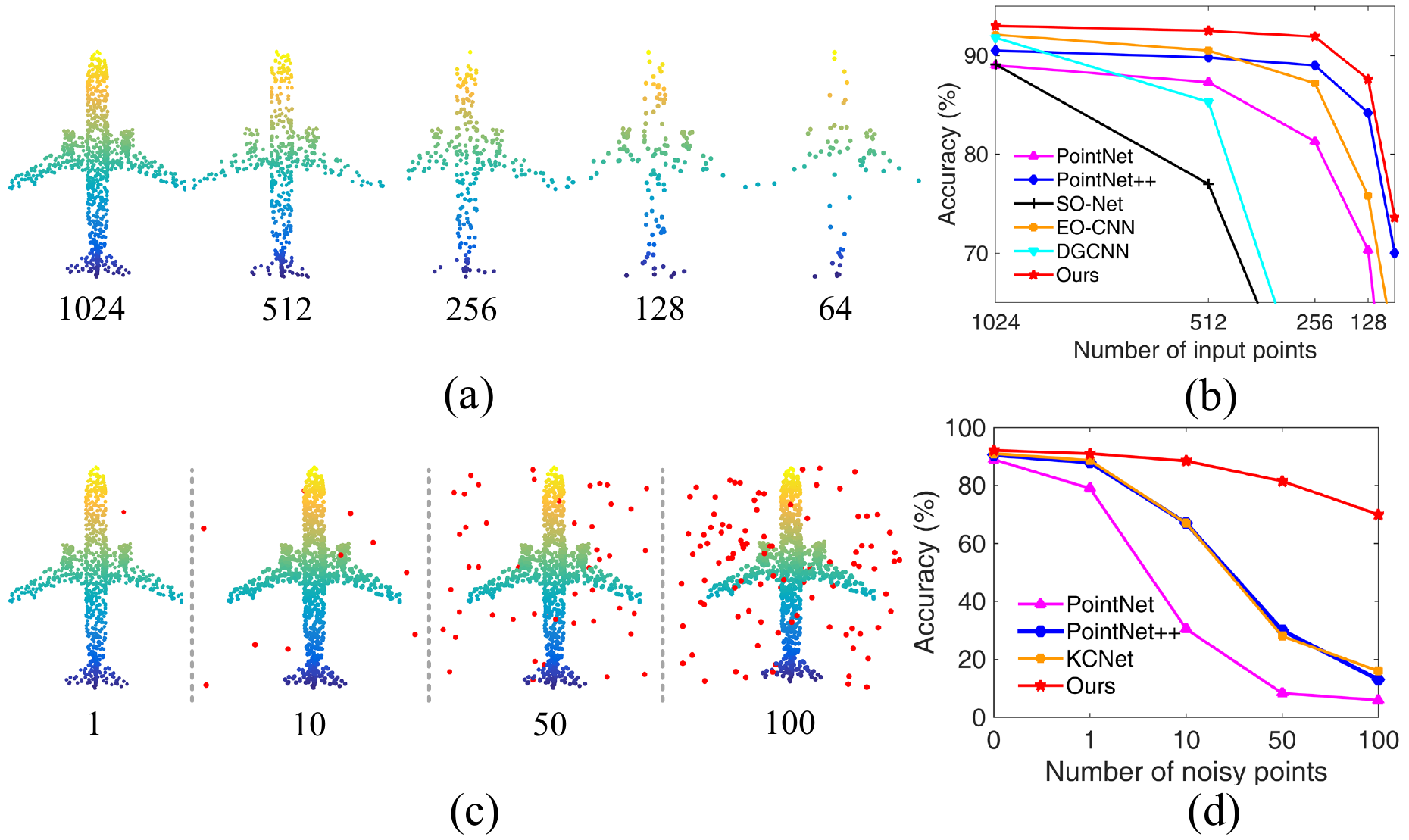}}
\caption{(a) Point cloud with different sampling densities. (b) Results of testing with sparser points. (c) Point cloud with some points being replaced with random noise (highlighted in \textcolor[rgb]{1.00,0.00,0.00}{red}). (d) Results of testing with noisy points.}
\label{fig7:dropout}
\end{figure}

\begin{table}[t]
  \centering
  \small
  \caption{The comparisons of different narrowness $k$ ($N_g=2$).}
  \begin{tabular}{cccc}
  \Xhline{0.8pt}
   narrowness $k$ & \#params & \#FLOPs/sample & acc. (\%) \\
  \Xhline{0.5pt}
   12 & 0.56M & 294M & 92.1 \\
   24 & 0.67M & 651M & \textbf{93.2} \\
   36 & 0.76M & 957M & 92.9 \\
   48 & 0.88M & 1310M & 92.7 \\
  \Xhline{0.8pt}
  \end{tabular}
  \label{Tab7:narrowness}
\end{table}

\begin{table}[t]
  \centering
  \small
  \caption{The comparisons of DensePoint applied in different stages of the classification network (Fig.~\ref{fig3:arch}(a)).}
  \begin{tabular}{cccc}
  \Xhline{0.8pt}
   model & 1\textsuperscript{st} stage & 2\textsuperscript{nd} stage & acc. (\%) \\
  \Xhline{0.5pt}
   $\widetilde{\text{A}}$ &  &  & 90.5 \\
   B & $\checkmark$ &  & 91.8 \\
   C &  & $\checkmark$ & 92.3 \\
   D & $\checkmark$ & $\checkmark$ & \textbf{93.2} \\
  \Xhline{0.8pt}
  \end{tabular}
  \label{Tab8:network_stage}
\end{table}

\vspace{10pt}
\noindent \textbf{Network stage to apply DensePoint.} \
To investigate the impact of contextual semantics at different levels on shape recognition, we also apply DensePoint with ePConv in different stages of the classification network (Fig.~\ref{fig3:arch}(a)). The results are summarized in Table~\ref{Tab8:network_stage}. The baseline (model $\widetilde{\text{A}}$) is set as the same as the model A in Table~\ref{Tab5:ablation} but equipped with ePConv for a fair comparison. One can see that DensePoint applied in the 1\textsuperscript{st} stage (model B) and the 2\textsuperscript{nd} stage (model C) can both bring a considerable boost, while the latter performs better. This indicates the higher-level contextual semantics in the 2\textsuperscript{nd} stage can result in a more powerful representation for shape recognition. Finally, with DensePoint in each stage for sufficiently contextual semantic information, the best result of 93.2\% can be reached.

\vspace{10pt}
\noindent \textbf{Robustness analysis.} \
The robustness of DensePoint on sampling density and random noise are shown in Fig.~\ref{fig7:dropout}. For the former, we use sparser points of 1024, 512, 256, 128 and 64, as the input to a model trained with 1024 points. Random input dropout is applied during training, for fair comparisons with PointNet~\cite{c1_pointnet}, PointNet++~\cite{c2_pointnet2}, SO-Net~\cite{c20}, PCNN~\cite{c16_eocnn} and DGCNN~\cite{c22-dgcnn}. Fig.~\ref{fig7:dropout}(b) shows that our model and PointNet++ perform better in this testing. Nevertheless, our model can obtain higher accuracy than PointNet++ at all densities. This indicates the densely contextual semantics of DensePoint, is much more effective than the traditional multi-scale information of PointNet++.

For the latter, as in KCNet~\cite{c9_kcnet}, we replace a certain number of randomly picked points with uniform noise ranging [-1.0, 1.0] during testing. The comparisons with PointNet, PointNet++ and KCNet are shown in Fig.~\ref{fig7:dropout}(d). Note that for this testing, our model is trained without any data augmentations to avoid confusion. As can be seen, our model is quite robust on random noise, while the others are vulnerable. This demonstrates the powerfulness of densely contextual semantics in DensePoint.

\vspace{10pt}
\noindent \textbf{Model complexity.} \
The comparisons of model complexity with the state of the arts are summarized in Table~\ref{Tab9:complexity}. As can be seen, our model is quite competitive and it can be the most efficient one with the network depth $L=6$ (accuracy 92.1\%). This shows its great potential for real-time applications, \textit{e.g.}, scene parsing in autonomous driving.

\vspace{10pt}
\noindent \textbf{Discussion of limitations.} \
(1) The density of local point clouds is not considered, which could lead to less effectiveness in greatly non-uniform distribution; (2) The importance of each level of context is not evaluated, which could lead to the difficulty in identifying very alike shapes.

\begin{table}[t]
  \centering
  \footnotesize
  \caption{The comparisons of model complexity (``-'': unknown).}
  \begin{tabular}{lcc}
  \Xhline{0.8pt}
  method & \#params & \#FLOPs/sample \\
  \Xhline{0.5pt}
  PointNet~\cite{c1_pointnet} & 3.50M & 440M \\
  PointNet++~\cite{c27} & 1.48M & 1684M \\
  DGCNN~\cite{c27} & 1.84M & 2767M \\
  SpecGCN~\cite{c27} & 2.05M & 1112M \\
  KCNet~\cite{c9_kcnet} & 0.90M & - \\
  PCNN~\cite{c27} & 8.20M & 294M \\
  PointCNN~\cite{c27} & 0.60M & 1581M \\
  \Xhline{0.5pt}
  Ours ($k=24, L=11$) & 0.67M & 651M \\
  Ours ($k=24, L=6$) & \textbf{0.53M} & \textbf{148M} \\
  \Xhline{0.8pt}
  \end{tabular}
  \label{Tab9:complexity}
\end{table}

\vspace{10pt}
\section{Conclusion}
\label{sec:conclusion}
In this work, DensePoint, a general architecture to learn densely contextual representation for efficient point cloud processing, has been proposed.
DensePoint extends regular grid CNN to irregular point configuration by an efficient generalized convolution operator.
Based on this operator, DensePoint develops a deep hierarchy and progressively aggregate multi-level and multi-scale semantics from it.
As a consequence, DensePoint can acquire sufficiently contextual information along with rich semantics in an organic manner, making it highly effective for implicit shape identification.
Extensive experiments on challenging benchmarks across four tasks, as well as thorough model analysis, have demonstrated that DensePoint achieves the state of the arts. In addition, DensePoint shows quite good robustness against noisy points, which could provide a promising direction for robust point cloud representation learning.

%

\newpage
{\small
\bibliographystyle{ieee_fullname}
\bibliography{egbib}

\begin{thebibliography}{10}\itemsep=-1pt

\bibitem{c16_eocnn}
Matan Atzmon, Haggai Maron, and Yaron Lipman.
\newblock Point convolutional neural networks by extension operators.
\newblock In {\em SIGGRAPH}, pages 1--14, 2018.

\bibitem{shapecontext}
Serge~J. Belongie, Jitendra Malik, and Jan Puzicha.
\newblock Shape matching and object recognition using shape contexts.
\newblock {\em {IEEE} Trans. Pattern Anal. Mach. Intell.}, 24(4):509--522,
  2002.

\bibitem{c37}
Yifan Feng, Zizhao Zhang, Xibin Zhao, Rongrong Ji, and Yue Gao.
\newblock {GVCNN}: Group-view convolutional neural networks for {3D} shape
  recognition.
\newblock In {\em CVPR}, pages 264--272, 2018.

\bibitem{c45}
Matheus Gadelha, Rui Wang, and Subhransu Maji.
\newblock Multiresolution tree networks for {3D} point cloud processing.
\newblock In {\em ECCV}, pages 105--122, 2018.

\bibitem{PCPNet}
Paul Guerrero, Yanir Kleiman, Maks Ovsjanikov, and Niloy~J. Mitra.
\newblock {PCPNet}: Learning local shape properties from raw point clouds.
\newblock {\em Comput. Graph. Forum}, 37(2):75--85, 2018.

\bibitem{multiview2}
Haiyun Guo, Jinqiao Wang, Yue Gao, Jianqiang Li, and Hanqing Lu.
\newblock Multi-view {3D} object retrieval with deep embedding network.
\newblock {\em {IEEE} Trans. Image Processing}, 25(12):5526--5537, 2016.

\bibitem{c49}
Zhizhong Han, Mingyang Shang, Zhenbao Liu, Chi{-}Man Vong, Yu{-}Shen Liu,
  Matthias Zwicker, Junwei Han, and C.~L.~Philip Chen.
\newblock {SeqViews2SeqLabels}: Learning {3D} global features via aggregating
  sequential views by {RNN} with attention.
\newblock {\em {IEEE} Trans. Image Processing}, 28(2):658--672, 2019.

\bibitem{conf_iccv_HeZRS15}
Kaiming He, Xiangyu Zhang, Shaoqing Ren, and Jian Sun.
\newblock Delving deep into rectifiers: {Surpassing} human-level performance on
  {ImageNet} classification.
\newblock In {\em ICCV}, pages 1026--1034, 2015.

\bibitem{resnet}
Kaiming He, Xiangyu Zhang, Shaoqing Ren, and Jian Sun.
\newblock Deep residual learning for image recognition.
\newblock In {\em CVPR}, pages 770--778, 2016.

\bibitem{triplet-loss}
Xinwei He, Yang Zhou, Zhichao Zhou, Song Bai, and Xiang Bai.
\newblock Triplet-{Center} loss for multi-view {3D} object retrieval.
\newblock In {\em CVPR}, pages 1945--1954, 2018.

\bibitem{c67}
Pedro Hermosilla, Tobias Ritschel, Pere{-}Pau V{\'{a}}zquez, Alvar Vinacua, and
  Timo Ropinski.
\newblock Monte carlo convolution for learning on non-uniformly sampled point
  clouds.
\newblock {\em {ACM} Trans. Graph.}, 37(6):235:1--235:12, 2018.

\bibitem{c17_pointwise}
Binh-Son Hua, Minh-Khoi Tran, and Sai-Kit Yeung.
\newblock Pointwise convolutional neural networks.
\newblock In {\em CVPR}, pages 974--993, 2018.

\bibitem{densenet}
Gao Huang, Zhuang Liu, Laurens van~der Maaten, and Kilian~Q. Weinberger.
\newblock Densely connected convolutional networks.
\newblock In {\em CVPR}, pages 2261--2269, 2017.

\bibitem{c31}
Haibin Huang, Evangelos Kalogerakis, Siddhartha Chaudhuri, Duygu Ceylan,
  Vladimir~G. Kim, and Ersin Yumer.
\newblock Learning local shape descriptors from part correspondences with
  multiview convolutional networks.
\newblock {\em {ACM} Trans. Graph.}, 37(1):6:1--6:14, 2018.

\bibitem{C28}
Qiangui Huang, Weiyue Wang, and Ulrich Neumann.
\newblock Recurrent slice networks for {3D} segmentation of point clouds.
\newblock In {\em CVPR}, pages 2626--2635, 2018.

\bibitem{Grouping}
Yani Ioannou, Duncan~P. Robertson, Roberto Cipolla, and Antonio Criminisi.
\newblock Deep roots: Improving {CNN} efficiency with hierarchical filter
  groups.
\newblock In {\em CVPR}, pages 5977--5986, 2017.

\bibitem{BN}
Sergey Ioffe and Christian Szegedy.
\newblock Batch normalization: Accelerating deep network training by reducing
  internal covariate shift.
\newblock In {\em ICML}, pages 448--456, 2015.

\bibitem{bcl}
Varun Jampani, Martin Kiefel, and Peter~V. Gehler.
\newblock Learning sparse high dimensional filters: Image filtering, dense crfs
  and bilateral neural networks.
\newblock In {\em CVPR}, pages 4452--4461, 2016.

\bibitem{c23}
Mingyang Jiang, Yiran Wu, and Cewu Lu.
\newblock {PointSIFT}: {A} {SIFT}-like network module for {3D} point cloud
  semantic segmentation.
\newblock {\em arXiv preprint arXiv:1807.00652}, 2018.

\bibitem{robotic}
David~Inkyu Kim and Gaurav~S. Sukhatme.
\newblock Semantic labeling of {3D} point clouds with object affordance for
  robot manipulation.
\newblock In {\em ICRA}, pages 5578--5584, 2014.

\bibitem{c26}
Roman Klokov and Victor~S. Lempitsky.
\newblock Escape from cells: Deep {Kd-Networks} for the recognition of {3D}
  point cloud models.
\newblock In {\em ICCV}, pages 863--872, 2017.

\bibitem{Alexnet}
Alex Krizhevsky, Ilya Sutskever, and Geoffrey~E. Hinton.
\newblock {ImageNet} classification with deep convolutional neural networks.
\newblock In {\em NeurIPS}, pages 1106--1114, 2012.

\bibitem{c8_superpoint}
Loic Landrieu and Martin Simonovsky.
\newblock Large-scale point cloud semantic segmentation with superpoint graphs.
\newblock In {\em CVPR}, pages 4558--4567, 2018.

\bibitem{c20}
Jiaxin Li, Ben~M. Chen, and Gim~Hee Lee.
\newblock {SO-Net}: Self-organizing network for point cloud analysis.
\newblock In {\em CVPR}, pages 9397--9406, 2018.

\bibitem{c30}
Ruoyu Li, Sheng Wang, Feiyun Zhu, and Junzhou Huang.
\newblock Adaptive graph convolutional neural networks.
\newblock In {\em AAAI}, pages 3546--3553, 2018.

\bibitem{c27}
Yangyan Li, Rui Bu, Mingchao Sun, and Baoquan Chen.
\newblock Point{CNN}: Convolution on {X}-transformed points.
\newblock In {\em NeurIPS}, pages 828--838, 2018.

\bibitem{liu2019rscnn}
Yongcheng Liu, Bin Fan, Shiming Xiang, and Chunhong Pan.
\newblock Relation-shape convolutional neural network for point cloud analysis.
\newblock In {\em CVPR}, pages 8895--8904, 2019.

\bibitem{vox2}
Daniel Maturana and Sebastian Scherer.
\newblock {VoxNet}: {A} {3D} convolutional neural network for real-time object
  recognition.
\newblock In {\em IROS}, pages 922--928, 2015.

\bibitem{relu}
Vinod Nair and Geoffrey~E Hinton.
\newblock Rectified linear units improve restricted boltzmann machines.
\newblock In {\em ICML}, pages 807--814, 2010.

\bibitem{c71}
Aude Oliva and Antonio Torralba.
\newblock The role of context in object recognition.
\newblock {\em Trends in Cognitive Sciences}, 11(12):520--527, 2007.

\bibitem{self-drive}
Charles~R. Qi, Wei Liu, Chenxia Wu, Hao Su, and Leonidas~J. Guibas.
\newblock Frustum {PointNets} for {3D} object detection from {RGB-D} data.
\newblock In {\em CVPR}, pages 918--927, 2018.

\bibitem{c1_pointnet}
Charles~R. Qi, Hao Su, Kaichun Mo, and Leonidas~J. Guibas.
\newblock {PointNet}: Deep learning on point sets for {3D} classification and
  segmentation.
\newblock In {\em CVPR}, pages 77--85, 2016.

\bibitem{c51}
Charles~Ruizhongtai Qi, Hao Su, Matthias Nie{\ss}ner, Angela Dai, Mengyuan Yan,
  and Leonidas~J. Guibas.
\newblock Volumetric and multi-view {CNN}s for object classification on {3D}
  data.
\newblock In {\em CVPR}, pages 5648--5656, 2016.

\bibitem{c2_pointnet2}
Charles~R. Qi, Li Yi, Hao su, and Leonidas~J. Guibas.
\newblock {PointNet++}: Deep hierarchical feature learning on point sets in a
  metric space.
\newblock In {\em NeurIPS}, pages 5099--5108, 2017.

\bibitem{c6_synccnn}
Siamak Ravanbakhsh, Jeff Schneider, and Barnabas Poczos.
\newblock Deep learning with sets and point clouds.
\newblock In {\em ICLR}, pages 1--12, 2017.

\bibitem{c53}
Gernot Riegler, Ali~Osman Ulusoy, and Andreas Geiger.
\newblock {OctNet}: Learning deep {3D} representations at high resolutions.
\newblock In {\em CVPR}, pages 6620--6629, 2017.

\bibitem{ImageNet}
Olga Russakovsky, Jia Deng, Hao Su, Jonathan Krause, Sanjeev Satheesh, Sean Ma,
  Zhiheng Huang, Andrej Karpathy, Aditya Khosla, Michael~S. Bernstein,
  Alexander~C. Berg, and Fei{-}Fei Li.
\newblock {ImageNet} large scale visual recognition challenge.
\newblock {\em International Journal of Computer Vision}, 115(3):211--252,
  2015.

\bibitem{PANORAMA}
Konstantinos Sfikas, Ioannis Pratikakis, and Theoharis Theoharis.
\newblock Ensemble of {PANORAMA}-based convolutional neural networks for {3D}
  model classification and retrieval.
\newblock {\em Computers {\&} Graphics}, 71:208--218, 2018.

\bibitem{c9_kcnet}
Yiru Shen, Chen Feng, Yaoqing Yang, and Dong Tian.
\newblock Mining point cloud local structures by kernel correlation and graph
  pooling.
\newblock In {\em CVPR}, pages 4548--4557, 2018.

\bibitem{c32}
Martin Simonovsky and Nikos Komodakis.
\newblock Dynamic edge-conditioned filters in convolutional neural networks on
  graphs.
\newblock In {\em CVPR}, pages 29--38, 2017.

\bibitem{VGG}
Karen Simonyan and Andrew Zisserman.
\newblock Very deep convolutional networks for large-scale image recognition.
\newblock In {\em ICLR}, pages 1--14, 2015.

\bibitem{dropout}
Nitish Srivastava, Geoffrey~E Hinton, Alex Krizhevsky, Ilya Sutskever, and
  Ruslan Salakhutdinov.
\newblock {Dropout}: {A} simple way to prevent neural networks from
  overfitting.
\newblock {\em Journal of Machine Learning Research.}, 15(1):1929--1958, 2014.

\bibitem{c10_splatnet}
Hang Su, Varun Jampani, Deqing Sun, Subhransu Maji, Evangelos Kalogerakis,
  Ming-Hsuan Yang, and Jan Kautz.
\newblock {SPLATNet}: Sparse lattice networks for point cloud processing.
\newblock In {\em CVPR}, pages 2530--2539, 2018.

\bibitem{multiview1}
Hang Su, Subhransu Maji, Evangelos Kalogerakis, and Erik~G. Learned{-}Miller.
\newblock Multi-view convolutional neural networks for {3D} shape recognition.
\newblock In {\em ICCV}, pages 945--953, 2015.

\bibitem{vox3}
Maxim Tatarchenko, Alexey Dosovitskiy, and Thomas Brox.
\newblock Octree generating networks: Efficient convolutional architectures for
  high-resolution {3D} outputs.
\newblock In {\em ICCV}, pages 2107--2115, 2017.

\bibitem{c19}
Gusi Te, Wei Hu, Amin Zheng, and Zongming Guo.
\newblock {RGCNN}: Regularized graph {CNN} for point cloud segmentation.
\newblock In {\em MM}, pages 746--754, 2018.

\bibitem{C3D}
Du Tran, Lubomir~D. Bourdev, Rob Fergus, Lorenzo Torresani, and Manohar Paluri.
\newblock Learning spatiotemporal features with {3D} convolutional networks.
\newblock In {\em ICCV}, pages 4489--4497, 2015.

\bibitem{selfatt}
Ashish Vaswani, Noam Shazeer, Niki Parmar, Jakob Uszkoreit, Llion Jones,
  Aidan~N. Gomez, Lukasz Kaiser, and Illia Polosukhin.
\newblock Attention is all you need.
\newblock In {\em NeurIPS}, pages 6000--6010, 2017.

\bibitem{c14_scn}
Chu Wang, Babak Samari, and Kaleem Siddiqi.
\newblock Local spectral graph convolution for point set feature learning.
\newblock In {\em ECCV}, pages 1--16, 2018.

\bibitem{c29}
Peng{-}Shuai Wang, Yang Liu, Yu{-}Xiao Guo, Chun{-}Yu Sun, and Xin Tong.
\newblock {O-CNN:} octree-based convolutional neural networks for {3D} shape
  analysis.
\newblock {\em {ACM} Trans. Graph.}, 36(4):72:1--72:11, 2017.

\bibitem{c70}
Peng{-}Shuai Wang, Chun{-}Yu Sun, Yang Liu, and Xin Tong.
\newblock Adaptive {O-CNN:} a patch-based deep representation of {3D} shapes.
\newblock {\em {ACM} Trans. Graph.}, 37(6):217:1--217:11, 2018.

\bibitem{c22-dgcnn}
Yue Wang, Yongbin Sun, Ziwei Liu, Sanjay~E. Sarma, Michael~M. Bronstein, and
  Justin~M. Solomon.
\newblock Dynamic graph {CNN} for learning on point clouds.
\newblock {\em {ACM} Trans. Graph.}, pages 1--13, 2019.

\bibitem{modelnet40}
Zhirong Wu, Shuran Song, Aditya Khosla, Fisher Yu, Linguang Zhang, Xiaoou Tang,
  and Jianxiong Xiao.
\newblock {3D ShapeNets}: {A} deep representation for volumetric shapes.
\newblock In {\em CVPR}, pages 1912--1920, 2015.

\bibitem{multiview3}
Jin Xie, Guoxian Dai, Fan Zhu, Edward~K. Wong, and Yi Fang.
\newblock {DeepShape}: Deep-learned shape descriptor for {3D} shape retrieval.
\newblock {\em {IEEE} Trans. Pattern Anal. Mach. Intell.}, 39(7):1335--1345,
  2017.

\bibitem{c7_attsp}
Saining Xie, Sainan Liu, Zeyu Chen, and Zhuowen Tu.
\newblock Attentional {ShapeContextNet} for point cloud recognition.
\newblock In {\em CVPR}, pages 4606--4615, 2018.

\bibitem{c21}
Yifan Xu, Tianqi Fan, Mingye Xu, Long Zeng, and Yu Qiao.
\newblock Spider{CNN}: Deep learning on point sets with parameterized
  convolutional filters.
\newblock In {\em ECCV}, pages 90--105, 2018.

\bibitem{c54}
Li Yi, Vladimir~G. Kim, Duygu Ceylan, I{-}Chao Shen, Mengyan Yan, Hao Su, Cewu
  Lu, Qixing Huang, Alla Sheffer, and Leonidas~J. Guibas.
\newblock A scalable active framework for region annotation in {3D} shape
  collections.
\newblock {\em {ACM} Trans. Graph.}, 35(6):210:1--210:12, 2016.

\bibitem{c3_synccnn}
Li Yi, Hao Su, Xingwen Guo, and Leonidas~J. Guibas.
\newblock {SyncSpecCNN}: Synchronized spectral {CNN} for {3D} shape
  segmentation.
\newblock In {\em CVPR}, pages 6584--6592, 2017.

\bibitem{c69}
Kangxue Yin, Hui Huang, Daniel Cohen{-}Or, and Hao~(Richard) Zhang.
\newblock {P2P-NET:} bidirectional point displacement net for shape transform.
\newblock {\em {ACM} Trans. Graph.}, 37(4):152:1--152:13, 2018.

\bibitem{c24}
Manzil Zaheer, Satwik Kottur, Siamak Ravanbakhsh, Barnab{\'{a}}s P{\'{o}}czos,
  Ruslan~R. Salakhutdinov, and Alexander~J. Smola.
\newblock Deep sets.
\newblock In {\em NeurIPS}, pages 3394--3404, 2017.

\bibitem{visualize}
Matthew~D. Zeiler and Rob Fergus.
\newblock Visualizing and understanding convolutional networks.
\newblock In {\em ECCV}, pages 818--833, 2014.

\end{thebibliography}
}

\newpage
\renewcommand{\thetable}{\Roman{table}}
\renewcommand\thesection{\Alph{section}}
\setcounter{section}{0}
\setcounter{table}{0}
{\Large \hspace{0.9cm} \textbf{Supplementary Material}}

\section{Outline}
\label{sec:Outline}
This supplementary material provides: (1) further investigations of the proposed DensePoint (Sec \ref{sec:experiments}); (2) more shape retrieval examples of DensePoint and some analysis (Sec \ref{sec:retrieval}); (3) network configuration details (Sec \ref{sec:configuration}). (4) training details (Sec \ref{sec:training}) 

\section{Further Investigations}
\label{sec:experiments}
In this section, we provide further investigations of DensePoint on four aspects. Specifically, the discussion of neighborhood method is presented in Sec \ref{subsec1}. The effect of dropout on ${\bm{\mathrm f}}_{\mathcal{N}(x)}$ in Eq. (\textcolor[rgb]{1.00,0.00,0.00}{4}) is analyzed in Sec \ref{subsec2}. The impact of network depth on classification performance is investigated in Sec \ref{subsec3}. The memory and runtime are summarized in Sec \ref{subsec4}. All the investigations are conducted on ModelNet40 dataset.

\subsection{Neighborhood Method}
\label{subsec1}
In the main paper, the local convolutional neighborhood $\mathcal{N}(x)$ in Eq. (\textcolor[rgb]{1.00,0.00,0.00}{1}) is set to be a spherical neighborhood, from which a fixed number of neighbors are randomly sampled for batch processing. We compare this strategy (Random-In-Sphere) with another typical one, \textit{i.e.}, k-nearest neighbor (k-NN). For a fair comparison, the models with these two strategies are configured with the same settings. Table \ref{Tab1:knn} summarizes the results.

As can be seen, the model with Random-In-Sphere performs better. We speculate that the model with k-NN will suffer from the distribution inhomogeneity of points. In this case, the contextual learning in DensePoint will be less effective, as the receptive fields will be confined to a local region with large density, which leads to ignoring those sparse points that are essential for recognizing the implicit shape. By contrast, Random-In-Sphere can have a better coverage of points even in the case of inhomogeneous distribution.

\subsection{Dropout on ${\bm{\mathrm f}}_{\mathcal{N}(x)}$ in Eq. (\textcolor[rgb]{1.00,0.00,0.00}{4})}
\label{subsec2}
The dropout technique can force the whole network to behave as an ensemble of a lot of subsets and reduce the risk of model overfitting. To analyze its effect on DensePoint, we apply it with different ratios on ${\bm{\mathrm f}}_{\mathcal{N}(x)}$ in Eq. (\textcolor[rgb]{1.00,0.00,0.00}{4}). The results are summarized in Table \ref{Tab2:dropout}. As can be seen, the best result of 93.2\% can be achieved with a dropout ratio of 20\%.

\subsection{Network Depth}
\label{subsec3}
We further explore the impact of the network depth (fully connected layers are not included) on classification performance. The results are summarized in Table \ref{Tab3:depth}. Surprisingly, a 6-layer network equipped with DensePoint can achieve an accuracy of 92.1\% with only 0.53M params and 148M FLOPs/sample. This even outperforms PointNet++ [\textcolor[rgb]{0.00,1.00,0.00}{33}] (accuracy 90.7\%, params 1.48M [\textcolor[rgb]{0.00,1.00,0.00}{26}], FLOPs/sample 1684M [\textcolor[rgb]{0.00,1.00,0.00}{26}]) by 15\% in error rate, whilst being one order of magnitude faster in terms of FLOPs/sample. We also observe that it is unnecessary to develop a very deep network (\textit{e.g.}, 23 layers) with DensePoint, as it increases complexity without bringing any gain. Eventually, the best result of 93.2\% can be reached with acceptable complexity by an 11-layer network.

\begin{table}[t]
  \centering
  \caption{The results (\%) of two neighborhood strategies. The number of neighbors is equally set in each layer of the two models.}
  \begin{tabular}{lc}
  \Xhline{0.8pt}
  neighborhood method & acc. \\
  \Xhline{0.5pt}
  k-NN  & 91.3 \\
  Random-In-Sphere & \textbf{93.2} \\
  \Xhline{0.8pt}
  \end{tabular}
  \label{Tab1:knn}
\end{table}

\begin{table}[t]
  \centering
  \caption{The results (\%) of dropout with different ratios applied on ${\bm{\mathrm f}}_{\mathcal{N}(x)}$ in Eq. (\textcolor[rgb]{1.00,0.00,0.00}{4}).}
  \begin{tabular}{lcccccc}
  \Xhline{0.8pt}
  ratio (\%) & 0 & 10 & 20 & 30 & 40 & 50 \\
  \Xhline{0.5pt}
  acc. & 92.9 & 92.8 & \textbf{93.2} & 93.0 & 92.8 & 92.5 \\
  \Xhline{0.8pt}
  \end{tabular}
  \label{Tab2:dropout}
\end{table}

\begin{table}[t]
  \centering
  \caption{The results (\%) of different network depths (fully connected layers are not included).}
  \begin{tabular}{cccc}
  \Xhline{0.8pt}
   \#layers & \#params & \#FLOPs/sample & acc. \\
  \Xhline{0.5pt}
   6 & 0.53M & 148M & 92.1 \\
   9 & 0.56M & 510M & 92.9 \\
   11 & 0.67M & 651M & \textbf{93.2} \\
   15 & 0.78M & 779M & 93.0 \\
   19 & 0.88M & 1222M & 92.7 \\
   23 & 1.03M & 1416M & 92.6 \\
  \Xhline{0.8pt}
  \end{tabular}
  \label{Tab3:depth}
\end{table}

\subsection{Memory and runtime}
\label{subsec4}
The memory and runtime of the proposed DensePoint are summarized in Table \ref{Tab:runtime}. As can be seen, the model ($L$=$11$) is competitive while another model ($L$=$6$) is the best one in terms of efficiency. Actually, the memory and training time issues in dense connection mode are greatly alleviated due to the shallow design of DensePoint and our highly-efficient implementation. Moreover, although extremely deep network could be unnecessary for 3D currently, in case of very deep DensePoint in the future, the technique of Shared Memory Allocations can be applied to achieve linear memory complexity.

\begin{table}[t]
  \centering
  \scriptsize
  \caption{\footnotesize Time and memory of classification network, where $k$ is network narrowness, $L$ is network depth. The statistics of all the models are summarized with batch size 16 on NVIDIA TITAN Xp, and time is the mean time of $1000$ tests. The compared models are tested using their available official codes.}
  \begin{tabular}{l|c|cc|cc}
  \Xhline{0.8pt}
  \multirow{2}*{method} & \multirow{2}*{\#points} & \multicolumn{2}{c|}{Time (ms)} & \multicolumn{2}{c}{Memory (GB)} \\
  \cline{3-6}
  & & training & test & training & test \\
  \Xhline{0.5pt}
  PointNet [\textcolor[rgb]{0.00,1.00,0.00}{31}] & 1024 & 55 & 22 & 1.318 & 0.469  \\
  PointNet++ [\textcolor[rgb]{0.00,1.00,0.00}{33}] & 1024 & 195 & 47 & 8.311 & 2.305  \\
  DGCNN [\textcolor[rgb]{0.00,1.00,0.00}{52}] & 1024 & 300 & 68 & 4.323 & 1.235  \\
  PointCNN [\textcolor[rgb]{0.00,1.00,0.00}{26}] & 1024 & 55 & 38 & 2.501 & 1.493  \\
  Ours ($k$=$24$, $L$=$11$) & 1024 & 21 & 10 & 3.745 & 1.228  \\
  Ours ($k$=$24$, $L$=$6$) & 1024 & 10 & 5 & 1.468 & 0.886  \\
  \Xhline{0.8pt}
  \end{tabular}
  \label{Tab:runtime}
  \vspace{-5pt}
\end{table} 

\section{Shape Retrieval}
\label{sec:retrieval}
\begin{figure*}[t]
\centerline{\includegraphics[width=16.5cm]{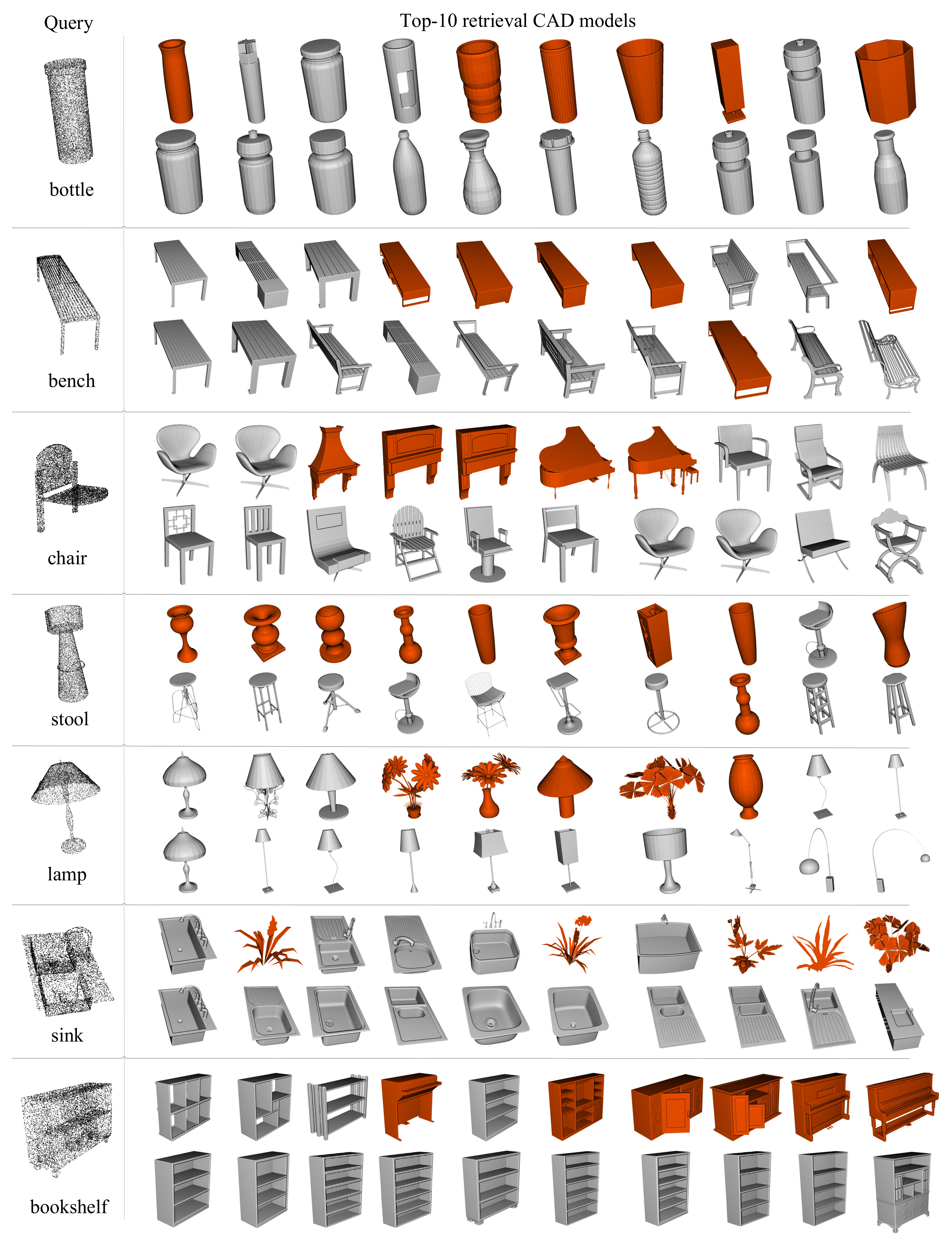}}
\caption{Retrieval examples on ModelNet40 dataset. Top-10 matches are shown for each query, with the 1\textsuperscript{st} line for PointNet [\textcolor[rgb]{0.00,1.00,0.00}{31}] and the 2\textsuperscript{nd} line for our DensePoint. The mistakes are highlighted in \textcolor[rgb]{1.00,0.00,0.00}{red}.}
\label{fig:retrieval}
\end{figure*}

In this section, we show more shape retrieval examples in Fig. \ref{fig:retrieval}. As can be seen, compared with PointNet [\textcolor[rgb]{0.00,1.00,0.00}{31}], our DensePoint obtains superior shape identification results. Specifically, PointNet is confused between the query ``bottle'' and the sample ``vase'' due to their similar shapes. Nevertheless, DensePoint with densely contextual semantics acquired can identify them accurately. We notice that DensePoint could also be confused for some very alike shapes, \textit{e.g.}, the query ``bench'' and the sample ``tv\_stand''. This could be improved by learning to weight multi-level contextual information instead of identically aggregating all levels of information. We leave it as future work. 

\section{Network Configuration Details}
\label{sec:configuration}
In this section, we present the configuration details of three networks on shape classification, shape part segmentation and normal estimation, respectively. For clearness, we describe the layer and corresponding setting format as follows:

\vspace{5pt}
\noindent \textbf{PPool:} [downsampling rate, neighborhood radius, \#number of neighbors, SLP$^{\phi}$(\#input channels, \#output channels)]. The global pooling is achieved by directly applying PConv to convolve all points.

\vspace{5pt}
\noindent \textbf{ePConv:} [neighborhood radius, \#number of neighbors, \\
\noindent SLP$^{\widetilde{\phi}}$(\#input channels, \#output channels, \#group number), \\
\noindent SLP$^{\psi}$(\#input channels, \#output channels), dropout ratio].

\vspace{5pt}
\noindent \textbf{FP (feature propagation layer):} MLP(\#channels, $\cdots$). Feature propagation layer [\textcolor[rgb]{0.00,1.00,0.00}{33}] is used for transforming the features that are concatenated from current interpolated layer and long-range connected layer. We employ a multi-layer perceptron (MLP) to implement this  transformation.

\vspace{5pt}
\noindent \textbf{FC (fully connected layer):} [(\#input channels, \#output channels), dropout ratio]. Note that the dropout technique is applied for all FC layers except for the last FC layer (used for prediction).

\vspace{5pt}
In addition, except for the last prediction layer, all layers (including the inside perceptrons) are followed with batch normalization and ReLU activator. The output shape is in the format of (\#feature dimension, \#number of points).

\subsection{Shape Classification Network}
The configuration details of shape classification network are presented in Table \ref{Tab4:cls}. The network has 14 layers in total, which comprises 3 PPools (the last one is global pooling layer) and 2 DensePoints (the 1\textsuperscript{st} one has 3 layers while the 2\textsuperscript{nd} one has 5 layers), followed by 3 FC layers.

\subsection{Shape Part Segmentation Network}
Table \ref{Tab5:seg} summarizes the configuration details of shape part segmentation network. As it shows, the network has 23 layers in total, which comprises 4 PPools, 3 DensePoints (4 layers, 6 layers and 3 layers in 2\textsuperscript{nd} stage, 3\textsuperscript{rd} stage and 4\textsuperscript{th} stage respectively) and 4 FP layers, followed by 2 FC layers. As in [\textcolor[rgb]{0.00,1.00,0.00}{31}, \textcolor[rgb]{0.00,1.00,0.00}{33}], we concatenate the one-hot encoding (16-d) of the object label to the last feature layer.

\subsection{Normal Estimation Network}
The normal estimation network is presented in Table \ref{Tab6:normal}. It is almost the same as the segmentation network, except for three aspects: (1) the input becomes 1024-d and the one-hot encoding becomes 40-d for ModelNet40 dataset; (2) the settings of some layers are slightly changed to be consistent with the 1024-d input; (3) the final output becomes 3-d for normal prediction.  As done in the segmentation network, we also concatenate the one-hot encoding (40-d) of the object label to the last feature layer.

\section{Training Details}
\label{sec:training}
Our DensePoint is implemented using Pytorch. The Adam optimization algorithm is employed for training, with a mini-batch size of $32$. The momentum for batch normalization starts with $0.9$ and decays with a rate of $0.5$ every $20$ epochs. The learning rate begins with $0.001$ and decays with a rate of $0.7$ every $20$ epochs. The weight is initialized using the techniques introduced by He \textit{et al}.~\cite{conf_iccv_HeZRS15}.

\begin{table*}[t]
  \centering
  \caption{The configuration details of shape part segmentation network. ``long-range'' indicates the long-range connections (see Fig. \textcolor[rgb]{1.00,0.00,0.00}{3}(b) in the main paper). $K$ is the number of classes.}
  \begin{threeparttable}
  \begin{tabular}{ccccc}
  \Xhline{0.8pt}
  stage & layer type & setting detail & output shape & long-range \\
  \Xhline{0.5pt}
  - & Input & - & (3, 2048) & FP$_{4}$ \\
  \Xhline{0.5pt}
  1 & PPool & [1/2, 0.1, 32, (3, 64)] & (64, 1024) & FP$_{3}$ \\
    \Xhline{0.5pt}
  \multirow{5}{*}{2} & PPool & [1/4, 0.2, 64, (64, 128)] & (128, 256) & \\
                  & ePConv & [0.3, 32, (128, 96, 2), (96, 24), 20\%] & (24, 256) & \\
                  & ePConv & [0.3, 32, (152, 96, 2), (96, 24), 20\%] & (24, 256) & \\
                  & ePConv & [0.3, 32, (176, 96, 2), (96, 24), 20\%] & (24, 256) & \\
                  & ePConv & [0.3, 32, (200, 96, 2), (96, 24), 20\%] & (24, 256) & \\
  \Xhline{0.5pt}
  \multicolumn{3}{r}{The output of DensePoint in 2\textsuperscript{nd} stage} & (224, 256) & FP$_{2}$ \\
  \Xhline{0.5pt}
  \multirow{7}{*}{3} & PPool & [1/4, 0.3, 32, (224, 192)] & (192, 64) & \\
                  & ePConv & [0.5, 16, (192, 96, 2), (96, 24), 20\%] & (24, 64) & \\
                  & ePConv & [0.5, 16, (216, 96, 2), (96, 24), 20\%] & (24, 64) & \\
                  & ePConv & [0.5, 16, (240, 96, 2), (96, 24), 20\%] & (24, 64) & \\
                  & ePConv & [0.5, 16, (264, 96, 2), (96, 24), 20\%] & (24, 64) & \\
                  & ePConv & [0.5, 16, (288, 96, 2), (96, 24), 20\%] & (24, 64) & \\
                  & ePConv & [0.5, 16, (312, 96, 2), (96, 24), 20\%] & (24, 64) & \\
  \Xhline{0.5pt}
  \multicolumn{3}{r}{The output of DensePoint in 3\textsuperscript{rd} stage} & (336, 64) & FP$_{1}$ \\
  \Xhline{0.5pt}
  \multirow{4}{*}{4} & PPool & [1/4, 0.8, 32, (336, 360)] & (360, 16) & \\
                  & ePConv & [0.8, 8, (360, 96, 2), (96, 24), 20\%] & (24, 16) & \\
                  & ePConv & [0.8, 8, (384, 96, 2), (96, 24), 20\%] & (24, 16) & \\
                  & ePConv & [0.8, 8, (408, 96, 2), (96, 24), 20\%] & (24, 16) & \\
  \Xhline{0.5pt}
  \multicolumn{3}{r}{The output of DensePoint in 4\textsuperscript{th} stage} & (432, 16) & \\
  \Xhline{0.5pt}
                  & FP$_{1}$ & (768, 512, 512) & (512, 64) & \\
                  & FP$_{2}$ & (736, 384, 384) & (384, 256) & \\
                  & FP$_{3}$ & (448, 256, 256) & (256, 1024) & \\
                  & FP$_{4}$ & (259, 128, 128) & (128, 2048) & \\
  \Xhline{0.5pt}
                  & FC & [(128+16$^{\textcolor[rgb]{1.00,0.00,0.00}{1}}$, 128), 50\%] & (128, 2048) & \\
                  & FC & [(128, $K$), -] $\rightarrow$ softmax & ($K$, 2048) & \\
  \Xhline{0.8pt}
  \end{tabular}
  \begin{tablenotes}
        \footnotesize
        \item[1] This is the one-hot encoding of the object label on ShapeNet part dataset.
  \end{tablenotes}
  \end{threeparttable}
  \label{Tab5:seg}
\end{table*}

\newpage

\newpage
\begin{table*}[t]
  \centering
  \caption{The configuration details of shape classification network. $K$ is the number of classes.}
  \begin{tabular}{cccc}
  \Xhline{0.8pt}
  stage & layer type & setting detail & output shape  \\
  \Xhline{0.5pt}
  - & Input & - & (3, 1024) \\
  \Xhline{0.5pt}
  \multirow{4}{*}{1} & PPool & [1/2, 0.25, 64, (3, 96)] & (96, 512) \\
                  & ePConv & [0.2, 32, (96, 96, 2), (96, 24), 20\%] & (24, 512) \\
                  & ePConv & [0.2, 32, (120, 96, 2), (96, 24), 20\%] & (24, 512) \\
                  & ePConv & [0.2, 32, (144, 96, 2), (96, 24), 20\%] & (24, 512) \\
  \Xhline{0.5pt}
  \multicolumn{3}{r}{The output of DensePoint in 1\textsuperscript{st} stage} & (168, 512) \\
  \Xhline{0.5pt}
  \multirow{6}{*}{2} & PPool & [1/4, 0.3, 64, (168, 144)] & (144, 128) \\
                  & ePConv & [0.4, 16, (144, 96, 2), (96, 24), 20\%] & (24, 128) \\
                  & ePConv & [0.4, 16, (168, 96, 2), (96, 24), 20\%] & (24, 128) \\
                  & ePConv & [0.4, 16, (192, 96, 2), (96, 24), 20\%] & (24, 128) \\
                  & ePConv & [0.4, 16, (216, 96, 2), (96, 24), 20\%] & (24, 128) \\
                  & ePConv & [0.4, 16, (240, 96, 2), (96, 24), 20\%] & (24, 128) \\
  \Xhline{0.5pt}
  \multicolumn{3}{r}{The output of DensePoint in 2\textsuperscript{nd} stage} & (264, 128) \\
  \Xhline{0.5pt}
  \multirow{4}{*}{3} & PPool & [-, -, 128, (264, 512)] & (512, ) \\
                  & FC & [(512, 512), 50\%] & (512, ) \\
                  & FC & [(512, 256), 50\%] & (256, ) \\
                  & FC & [(256, $K$), -] $\rightarrow$ softmax & ($K$, ) \\
  \Xhline{0.8pt}
  \end{tabular}
  \label{Tab4:cls}
\end{table*}

\begin{table*}[t]
  \centering
  \caption{The configuration details of normal estimation network. ``long-range'' indicates the long-range connections (see Fig. \textcolor[rgb]{1.00,0.00,0.00}{3}(b) in the main paper).}
  \begin{threeparttable}
  \begin{tabular}{ccccc}
  \Xhline{0.8pt}
  stage & layer type & setting detail & output shape & long-range \\
  \Xhline{0.5pt}
  - & Input & - & (3, 1024) & FP$_{4}$ \\
  \Xhline{0.5pt}
  1 & PPool & [1, 0.2, 32, (3, 64)] & (64, 1024) & FP$_{3}$ \\
    \Xhline{0.5pt}
  \multirow{5}{*}{2} & PPool & [1/4, 0.2, 32, (64, 128)] & (128, 256) & \\
                  & ePConv & [0.3, 32, (128, 96, 2), (96, 24), 20\%] & (24, 256) & \\
                  & ePConv & [0.3, 32, (152, 96, 2), (96, 24), 20\%] & (24, 256) & \\
                  & ePConv & [0.3, 32, (176, 96, 2), (96, 24), 20\%] & (24, 256) & \\
                  & ePConv & [0.3, 32, (200, 96, 2), (96, 24), 20\%] & (24, 256) & \\
  \Xhline{0.5pt}
  \multicolumn{3}{r}{The output of DensePoint in 2\textsuperscript{nd} stage} & (224, 256) & FP$_{2}$ \\
  \Xhline{0.5pt}
  \multirow{7}{*}{3} & PPool & [1/4, 0.3, 32, (224, 192)] & (192, 64) & \\
                  & ePConv & [0.5, 16, (192, 96, 2), (96, 24), 20\%] & (24, 64) & \\
                  & ePConv & [0.5, 16, (216, 96, 2), (96, 24), 20\%] & (24, 64) & \\
                  & ePConv & [0.5, 16, (240, 96, 2), (96, 24), 20\%] & (24, 64) & \\
                  & ePConv & [0.5, 16, (264, 96, 2), (96, 24), 20\%] & (24, 64) & \\
                  & ePConv & [0.5, 16, (288, 96, 2), (96, 24), 20\%] & (24, 64) & \\
                  & ePConv & [0.5, 16, (312, 96, 2), (96, 24), 20\%] & (24, 64) & \\
  \Xhline{0.5pt}
  \multicolumn{3}{r}{The output of DensePoint in 3\textsuperscript{rd} stage} & (336, 64) & FP$_{1}$ \\
  \Xhline{0.5pt}
  \multirow{4}{*}{4} & PPool & [1/4, 0.8, 32, (336, 360)] & (360, 16) & \\
                  & ePConv & [0.8, 8, (360, 96, 2), (96, 24), 20\%] & (24, 16) & \\
                  & ePConv & [0.8, 8, (384, 96, 2), (96, 24), 20\%] & (24, 16) & \\
                  & ePConv & [0.8, 8, (408, 96, 2), (96, 24), 20\%] & (24, 16) & \\
  \Xhline{0.5pt}
  \multicolumn{3}{r}{The output of DensePoint in 4\textsuperscript{th} stage} & (432, 16) & \\
  \Xhline{0.5pt}
                  & FP$_{1}$ & (768, 512, 512) & (512, 64) & \\
                  & FP$_{2}$ & (736, 384, 384) & (384, 256) & \\
                  & FP$_{3}$ & (448, 256, 256) & (256, 1024) & \\
                  & FP$_{4}$ & (259, 128, 128) & (128, 1024) & \\
  \Xhline{0.5pt}
                  & FC & [(128+40$^{\textcolor[rgb]{1.00,0.00,0.00}{2}}$, 128), 50\%] & (128, 1024) & \\
                  & FC & [(128, 3), -] & (3, 1024) & \\
  \Xhline{0.8pt}
  \end{tabular}
  \begin{tablenotes}
        \footnotesize
        \item[2] This is the one-hot encoding of the object label on ModelNet40 dataset.
  \end{tablenotes}
  \end{threeparttable}
  \label{Tab6:normal}
\end{table*}

\end{document}